# Altair-Nidhogg: Which Model-Merging Theories Hold for a One-Pass Decision Encoder? Soups, Task Arithmetic, TIES and DARE Against Joint Training and Prediction Ensembling

*A test of published merging recipes and of the explanations given for them, on seeds, specialists, joint fine-tunes and held-out shifts of one non-autoregressive decision model*

Mohammed Nayeem Khan
mohammed.nayeem.khan@intel.com
mnknayeem@gmail.com

Navin Kumar R
navin.kumar.r@intel.com

Akhil P
akhil.p@intel.com

Intel Corporation

**Abstract**

Laya is a small model that answers a multiple-choice question in one pass: each option has a name and a written description. A new skill costs a training run, so we test whether published recipes for merging fine-tuned models (weight averaging, task arithmetic, TIES, DARE) work for it, and whether the theories behind them hold. All 156 models start from one released checkpoint: four plain fine-tunes, five specialists (each adds one kind of training row), joint fine-tunes on all rows, and prediction ensembles. Items are held-out, synthetic and LLM-labelled. In-family, each skill is scored on new items of its own kind, and "kept" is the share of a specialist's gain over the plain fine-tune. Averaging four specialists keeps 52% [38, 64] with +1.7 points of general accuracy; adding the full vectors keeps 42% [3, 63] with -9.0; TIES, which trims and resolves sign conflicts, keeps 64% [50, 76] with +1.5; joint fine-tunes, which see all the rows, keep 120% [104, 162] and 116% [104, 149]. DARE does not repair the full sum (-0.4 points), and ensembles of the models' predictions keep 26% to 54% depending on how they are combined. Swapping the seed of one specialist moves the average by up to 8 points (23 for the full sum). Of 10 predictions fixed in advance, 1 held, 1 held partly, 7 failed and 1 could not be tested. Out-of-family (benchmark questions, code-mixed tickets), the four-way average, joint models and ensemble stay within 2.6 points of the plain fine-tune, but full-vector sums lose up to 7.7. Labels are LLM-made, data synthetic, one base model.

**Key findings (every number recomputed from the frozen evaluations)**

1. Soups of seeds are free but not better: the four-seed v1 soup differs from its members by +0.3 points [-0.6, +1.3] on 743 items, and the v1-seed-1 line has a barrier of +0.7 points [+0.1, +1.5].
2. Halfway to a specialist keeps name-randomised 77%, HR-schema 83%, injection-hardened 33%, long-context 75% of the gain (in-family).
3. Averaging the four specialists (lambda 0.25) keeps 52% [38, 64] of their gains with +1.7 [+0.3, +3.4] points of general accuracy; full vectors (lambda 1) keep 42% [3, 63] with -9.0 [-11.7, -6.2].
4. Joint fine-tuning on all rows beats every merge: it keeps 120% [104, 162] and 116% [104, 149] of the four gains, +68 points [+51, +108], +65 points [+47, +98] above the four-way average, with general accuracy +0.3 [-1.3, +2.2], +0.9 [-0.7, +2.7] points from v1.
5. Prediction ensembling is not a free upper bound either: across six ensembles (three ways of combining, with and without v1) the gain kept runs from 26% to 54%, against 52% for the weight average. Only picking the most confident member matches the average, at four or five forward passes per call.
6. Random drop-and-rescale does not repair the full sum: best change in general accuracy -0.4 points against the plain sum; without the rescale at p 0.9 the gain kept falls to 22%.
7. The headline merges are seed-sensitive: replacing the rename vector by its other seeds changes the mean gain kept by up to 8 points at lambda 0.25 and 23 points at lambda 1.
8. Interference is moderate: a specialist change is 1.0-1.1 times a seed difference in size, different specialists have cosine 0.45 on average, and the large coordinates agree in sign 88% of the time (chance 50%).
9. TIES helps where the plain sum fails: at its usual setting (keep 20%, lambda 1) it keeps 64% [50, 76] with +1.5 [-0.3, +3.2] points of general accuracy; the best setting of the reduced grid (TIES from v1, keep 20%, lambda 1.5, picked on the test items) keeps 76%. See Table 13.
10. Moving only the head, scorer and act head to the specialist keeps 0%, 0%, 0%, -1% of the four gains (6% of the parameters).
11. Scorecard: of 10 predictions written down before the new runs, 2 were confirmed (or partly), 7 refuted and 1 untestable (Table 20).
12. Out-of-family, the four-way average, the joint models and the ensemble stay within 2.6 points of v1 on public benchmark questions and code-mixed tickets; the full-vector sums, TIES and DARE at lambda 1 lose up to 7.7 points on code-mixed tickets.

## 1 Introduction

Laya is a small model that answers a multiple-choice question about a text in one pass: the caller gives each option a name and a written description. A Laya checkpoint is small enough to retrain in about five minutes on one consumer GPU, and the project we work in has done that many times: a general fine-tune, then variants that add one kind of training row to fix one kind of failure. Each variant fixes its own failure and costs a new training run, a new test of everything else, and a new file. Weight-space merging promises a cheaper path. If the variants all start from the same released weights, perhaps their changes can be averaged, added or interpolated and the skills will come along, with no data and no trainer. This paper asks whether that works for a typed-decision encoder, and where it stops working.

We have a collection that makes the question testable. All fine-tunes start from the same released Laya weights and train on the same **2,818** generated rows. There are four seeds of one plain fine-tune ("v1"), five specialists that each add one ingredient to the training data or recipe (rewritten option names, an HR label schema, injected instructions, long padded inputs, or option shuffling switched off), and two extra seeds of two of the specialists. That gives us four kinds of pairs: same recipe with a different seed, different recipe, base weights against fine-tune, and specialist against specialist. We also have a specific, measurable skill for every specialist, so we can say what a merge keeps rather than only whether accuracy stayed flat.

We ask ten questions. The first six are about the methods themselves. **Q1**: does averaging seeds of the same recipe (a "soup") help, and is there an accuracy barrier on the straight line between two seeds? **Q2**: along the line from the released weights to v1, and from v1 to each specialist, how do the specialist skill and the general accuracy change, and is there any *alpha* where we get the skill without the loss? **Q3**: when several specialist changes are added or averaged (task arithmetic), what fraction of each gain survives, and how does that depend on the scale *lambda*? **Q4**: why does it work or fail: how large are the changes, how aligned are they, and how many coordinates conflict in sign? **Q5**: does trimming and sign election (TIES-style merging) repair the failures? **Q6**: which layers carry a specialist change, so that moving only some of them would be enough? The other four test the explanations that the merging literature gives. **Q7**: how does a merge compare with the thing it replaces, a joint fine-tune on all the data, and with averaging predictions instead of weights? **Q8**: does random drop-and-rescale (DARE) remove the interference between specialist vectors? **Q9**: do the headline merges depend on one seed, and do vectors transfer to another seed? **Q10**: do merges keep or lose ability on distributions that no model was trained on?

Everything is measured on the same held-out sets: 743 English items from six sets for general accuracy, 322 multilingual items, and five specialist-specific suites. We report Wilson intervals for accuracies and paired item bootstraps for differences. About **156** models (fine-tunes, merges, a joint fine-tune and prediction ensembles) were evaluated. The merged models were assembled in memory and never saved to disk, so every number below is recomputed when this document is built.

**What is and is not new.** The merging methods are not ours: soups, task arithmetic, TIES and DARE are published (Section References and related work), and so are the theories that explain them (task vectors as multitask gradients, redundancy of fine-tuning changes, orthogonality of task vectors, barriers that grow with distance, mergeability that follows the base model's prior knowledge). We found no published test of those theories on a one-pass typed-decision model like Laya. So the paper is not a new method; it is a careful test, on one model family, of which merging theories hold, against the two baselines that matter: a joint fine-tune and an ensemble of predictions. Each theory was turned into a numerical prediction before the new experiments were run (Section 16), and we report the ones that failed. We do not claim that the findings carry over to larger models, other bases or other kinds of specialist.

## 2 Laya, the data and the fine-tunes

### 2.1 The model

Laya is a non-autoregressive decision model: it never writes text. Each question comes as a header (the question and its options, every option introduced by a mask token) followed by the case text. A ModernBERT-large encoder (28 layers, hidden size 1024, rotary positions) reads the whole sequence once. The hidden state at each mask token goes through a small two-layer self-attention head that lets the options see each other; a scorer then gives one logit per option and a separate act head gives the probability that the model should answer at all. A temperature stored in the checkpoint turns logits into probabilities. In total there are **421.3** million parameters in **206** tensors: 394.8 million in the encoder (embeddings included) and 26.5 million in the head, scorer and act head. The checkpoint is stored in half precision, which matters for merging: in v1, 16% of all coordinates are bit-for-bit identical to the released weights after fine-tuning, because small updates round away. For the specialists the figures are 15% (HR), 15% (injection) and 15% (long-context).

### 2.2 The data and the labels

All cases are synthetic. For each category a large language model (Qwen3.8-27B) is asked, in batches of ten, for cases that a careful reader would label with a stated target label. A second, *blind* pass gives the same model the case without the target and keeps the case only if the answer agrees. The categories cover negation, multilingual text, 12-way choices, urgency scores, churn, injected instructions, JSON-serialised tickets, hardware failures and long noisy documents. To check the labels with a different model, gemma4-31b labelled every held-out item independently; it agreed with the stored label on probe 97%, sent 99%, json 97%, hw 95%, many 94%, ml 97%, urg 95%, hr 97%. The labels are LLM-made, not human, and the agreement shows consistency between two models, not correctness. Held-out sets are never used for training. The six English sets that make up "general accuracy" are probe (224), sent (106), json (58), hw (82), many (195), urg (78), 743 items in all; the multilingual set (322 items) is reported separately because it measures a different kind of text; the training rows do include 399 monolingual non-English rows, so it is a language shift inside the family, not an out-of-family test.

### 2.3 The fine-tunes

All models were trained with Laya's own trainer from the same released checkpoint: 4 epochs, encoder learning rate 2.5e-5, head learning rate 1e-4, micro-batch 8, bf16 autocast, and options shuffled for choice questions (except the no-shuffle variant). The specialists differ only in the rows or in one switch, as listed in Table 1. Because every model starts from the released weights and the rows overlap, each specialist is v1 plus a small extra change; the question for merging is whether those extra changes can be moved around.

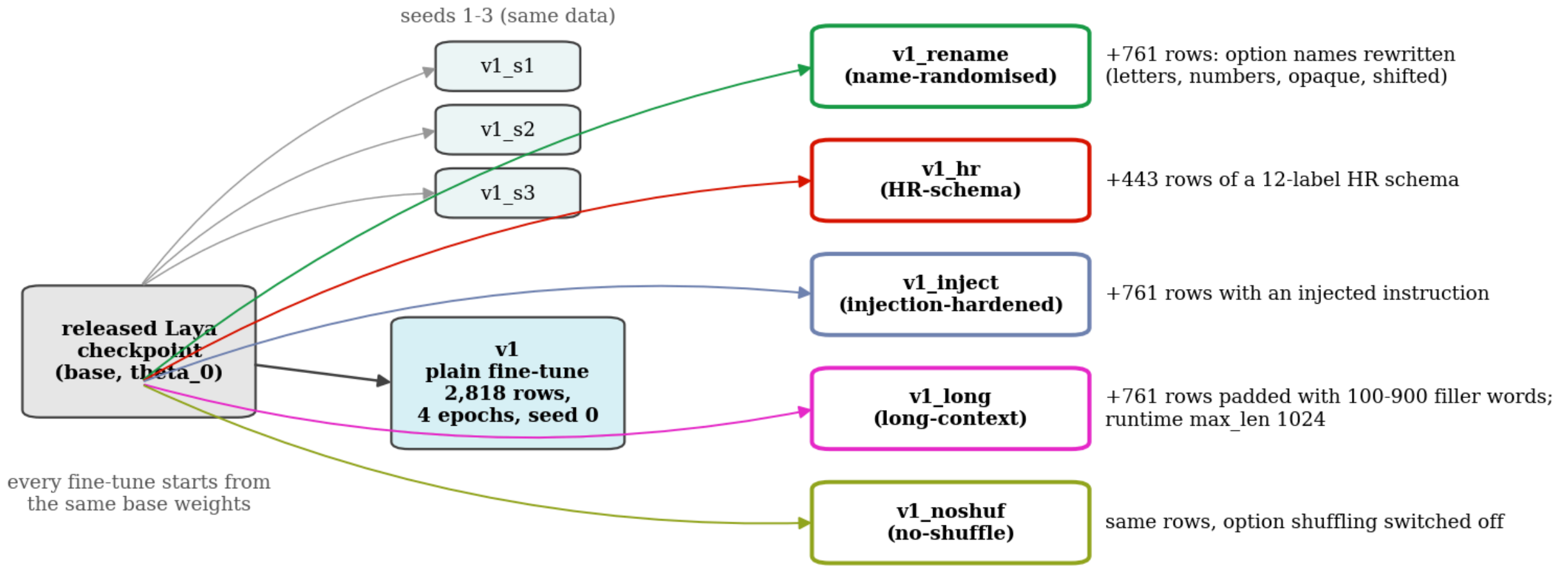

**Figure 1.** The fine-tunes used here. Every model starts from the released Laya weights. v1 is the plain fine-tune (four seeds); each specialist adds one ingredient to the same rows. Row counts are read from the training files.

**Table 1.** The fine-tunes and what each was trained on. Rows are the training-set size; the last column says how the specialist skill is tested.

| Model | Change to the recipe | Rows | Specialist skill | Held-out test |
|---|---|---|---|---|
| v1 (4 seeds) | plain recipe | 2,818 | reference | - |
| name-randomised (3 seeds) | a seeded 30% copy of the choice rows with option names rewritten (letters, numbers, 3-letter strings, or names shifted onto the wrong descriptions) | 3,579 | swap-name accuracy | held-out items, rename scheme (swap) not used in training |
| HR-schema | adds one 12-label HR request schema | 3,261 | HR-test accuracy | held-out HR requests (schema seen in training) |
| injection-hardened | 30% copy with an instruction naming a wrong label added before or after the text (5 templates) | 3,579 | accuracy under held-out injections | held-out templates (not in training) |

| Model | Change to the recipe | Rows | Specialist skill | Held-out test |
|---|---|---|---|---|
| long-context | 30% copy padded with 100-900 filler words; runtime max_len raised to 1024 | 3,579 | accuracy with held-out filler | held-out filler text (not in training) |
| no-shuffle (3 seeds) | the v1 rows, option shuffling switched off | 2,818 | order-stable decisions | rows from the same sets; permutations never seen in training |

Two limits follow from this design and they shape what we can conclude. First, every specialist shares the base and about 79-86% of its training rows with v1, so the specialist *differences* (specialist minus v1) are the interesting object, and they are of the same order as the difference between two seeds (Section 8). Second, three of the five recipes have only one seed, so for them we cannot separate a recipe effect from seed noise except through the seeds of the other recipes.

## 3 Merge operators and evaluation protocol

### 3.1 Merge operators

Let $\theta_0$ be the released weights and $\theta_m$ the weights of fine-tune $m$. A **soup** is the uniform mean of $k$ fine-tunes of one recipe. A **line** is $(1-\alpha)\,\theta_a + \alpha\,\theta_b$, with $\alpha$ outside [0,1] meaning extrapolation. **Task arithmetic** adds task vectors $\tau_m = \theta_m - \theta_{root}$ to a root: $\theta_{root} + \lambda \sum_{\mathrm{m}} \tau_m$. We root at v1 (so a vector is the specialist's extra change) and, for comparison, at the released weights. **TIES-style** merging keeps, per weight tensor, the top fraction $d$ of each vector by magnitude, elects the sign with the larger total mass per coordinate, and averages only the surviving entries that agree with it, then scales by $\lambda$. The scale matters in every one of these: $\lambda$=1/$k$ gives the mean of the specialists, $\lambda$=1 gives the full sum.

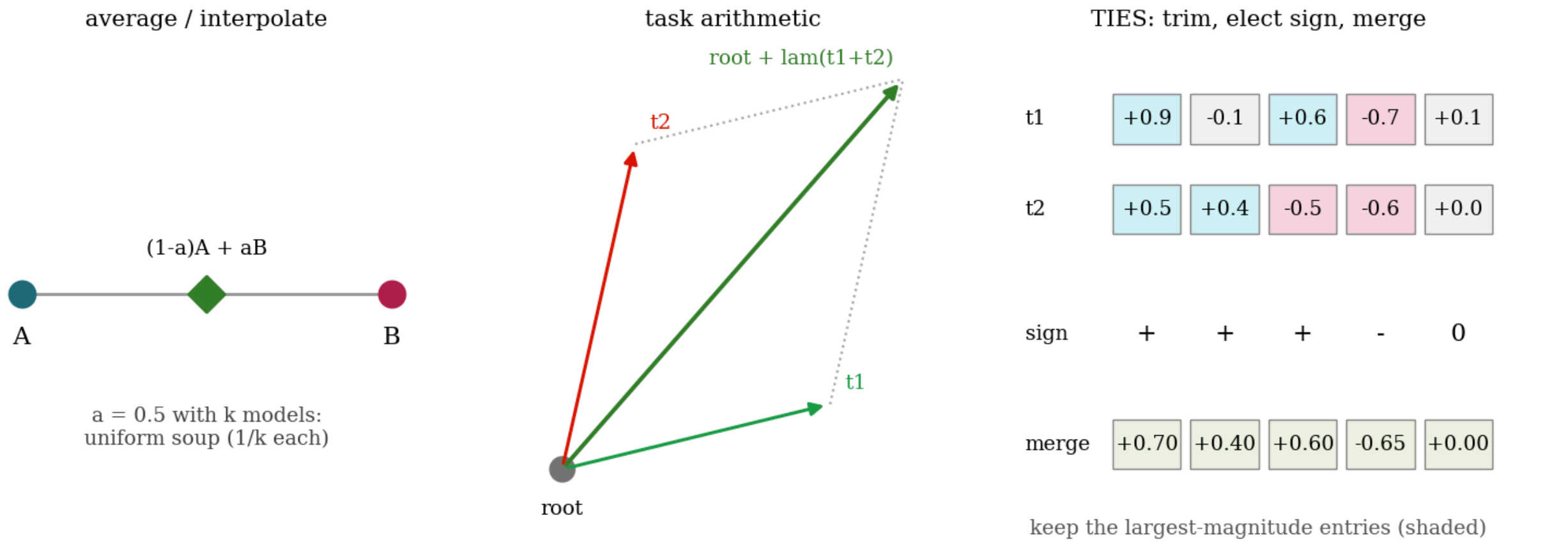

**Figure 2.** The three kinds of merge. Left: a straight line between two models, and the uniform average of k models. Middle: task arithmetic adds scaled differences (task vectors) to a root model. Right: TIES keeps only the largest entries of each vector, elects a sign per coordinate, and averages the entries that agree with it (toy numbers).

### 3.2 Protocol

Every merged model is built in 32-bit floating point on the CPU, loaded into the one resident Laya agent, and evaluated with exactly the code path used for the fine-tunes. The tokenizer, configuration and stored temperature are those of v1, so a merged model never gets its own recalibration; ECE therefore reflects what a user who merges and deploys would get. Merged weights are not written to disk. Sanity check: the 50/50 average of v1 and its seed 1 was built twice, once as a soup and once as a line; the two are the same weights, and their predictions differ on **0.00%** of all decisions, which is the run-to-run noise floor of the evaluation (it is non-zero because batched half-precision kernels are not bit-reproducible).

**Table 2.** The evaluation suite applied to every merged model. All sets are held out; none of the merged models was trained on any of them.

| Measure | Items | n | Interval |
|---|---|---|---|
| general accuracy | probe 224, sentiment 106, JSON 58, hardware 82, 12-way 195, urgency 78 | 743 | Wilson 95%; paired bootstrap for differences |
| multilingual accuracy | held-out multilingual items (English-root base; the training rows hold some multilingual text) | 322 | Wilson 95% |
| swap-name accuracy | option names permuted onto the wrong descriptions, over all multi-option questions | 951 | Wilson 95% |
| HR-test accuracy | unseen 12-label HR requests | 273 | Wilson 95% |
| accuracy under injection | held-out injection templates naming a wrong label | 106 | Wilson 95% |
| accuracy with filler | the sentiment set padded with 100-800 filler words at 6 length and position settings; per-item mean over settings | 106 items | Wilson 95% on item means |
| order stability | share of items whose decision is the same under every option permutation | 951 | Wilson 95% |

### 3.3 Measures

**Skill gain kept** ("retention") for a specialist $s$ and merged model $M$ is [skill($M$) − skill(v1)] / [skill($s$) − skill(v1)]: 100% means the merge matches the specialist, 0% means it matches v1, and negative values mean it is worse than v1. Its interval is a bootstrap over items, resampling the three models on their own items. Retention is only meaningful when the specialist's own gain is clearly non-zero; for the no-shuffle variant it is not (its order stability is lower than v1's), so we report its skill change directly. **Barrier** along a line is the mean of the two endpoint accuracies minus the lowest interior accuracy, so a positive number is a dip; its interval comes from a bootstrap over items. **Decisions flipped** is the share of all 1,065 test decisions (all seven sets) whose top option differs from v1's. ECE uses 15 equal-width bins.

### 3.4 What training and test share

Every test item is held out: none of the seven general sets or the specialist suites appears in any training file (the trainer's own overlap check found none for the probe set). But most of the evaluation is **in-family**. The general sets come from the same generator, the same labeling model and the same categories as the training rows, and each specialist skill is measured on items of the kind that specialist was trained on, with new items, new injection templates or new filler text. A merge that keeps a skill by this measure has kept it for that family. The **out-of-family** measures are the two shifts of Section 15: public benchmark questions, and tickets written in code-mixed languages. The multilingual set sits between the two: the training rows include some non-English text, so we treat it as a language shift inside the family. The abstract states which kind each headline number is.

## 4 Seed soups

The cheapest merge is the average of seeds of one recipe: the models differ only in training noise, so if the weights are close to each other in a good basin, their mean should be at least as good as any of them. The v1 family gives 11 soups (all subsets of size two, three and four of four seeds); the name-randomised and no-shuffle families have three seeds each.

For the v1 family the pairwise soups have an interval that contains zero, so we cannot show any gain over a single seed (Table 3: +0.4 points [-0.2, +0.9]); the four-seed soup is +0.3 points from its members [-0.6, +1.3]. The more informative change is in behaviour rather than accuracy: two v1 seeds disagree on 9.5% of all test decisions, a pairwise soup differs from each of its members on 5.2% and the four-seed soup on 6.5%. A soup is thus a consensus model that stays close to every member without being identical to any of them. Soups of the three-seed specialist families that were not evaluated within the compute budget are absent from the table; the pairs that were evaluated are included.

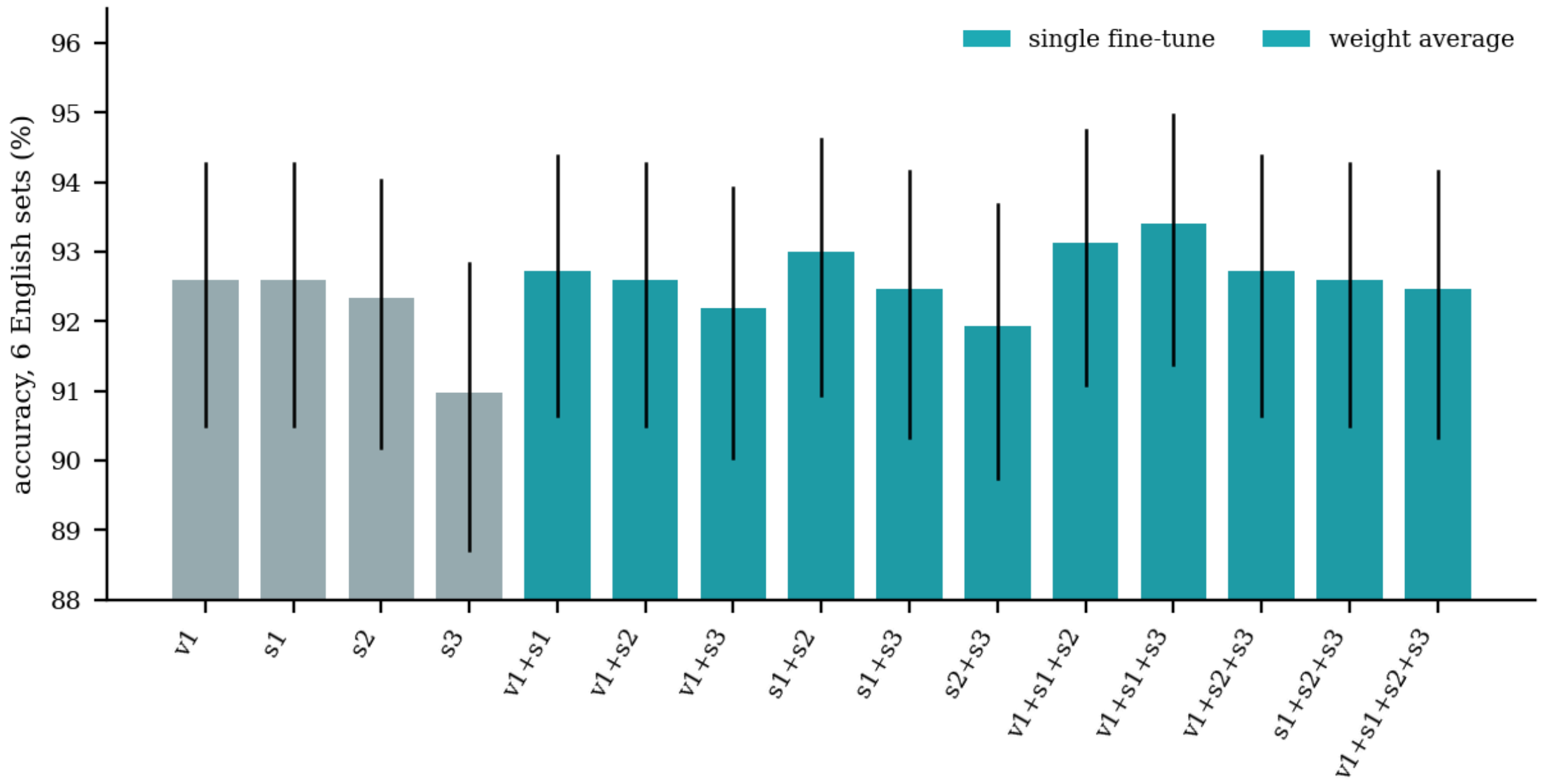

**Figure 3.** General accuracy (six English sets, n=743) of the four v1 seeds and of every uniform average of two, three and four of them, with Wilson intervals. Grey: single fine-tunes. Blue: weight averages.

**Table 3.** Uniform soups of seeds of one recipe. "Members" and "soups" are means over all soups of that size; the gain is each soup minus the mean of its own members, averaged over soups, with a bootstrap interval over items. Best member is the highest single-model value in the family.

| Recipe | k | Soups | Measure | Mean of members (%) | Mean of soups (%) | Gain, points [95% CI] | Best member (%) |
|---|---|---|---|---|---|---|---|
| v1 | 2 | 6 | general accuracy | 92.1 | 92.5 | +0.4 [-0.2, +0.9] | 92.6 |
| v1 | 2 | 6 | multilingual accuracy | 74.7 | 75.2 | +0.5 [-0.7, +1.8] | 77.0 |
| v1 | 3 | 4 | general accuracy | 92.1 | 93.0 | +0.8 [+0.0, +1.6] | 92.6 |
| v1 | 3 | 4 | multilingual accuracy | 74.7 | 75.2 | +0.5 [-1.1, +2.0] | 77.0 |
| v1 | 4 | 1 | general accuracy | 92.1 | 92.5 | +0.3 [-0.6, +1.3] | 92.6 |
| v1 | 4 | 1 | multilingual accuracy | 74.7 | 75.2 | +0.5 [-1.9, +2.9] | 77.0 |
| rename | 2 | 3 | general accuracy | 92.6 | 93.7 | +1.1 [+0.6, +1.7] | 93.0 |
| rename | 2 | 3 | swap-name accuracy | 83.4 | 85.0 | +1.6 [+0.8, +2.5] | 85.9 |

With 743 items, a one-point difference is inside the interval of every single model (about ±1 point), so the reading we can defend is "a soup of seeds costs nothing and does not help measurably"; the soup's value is that one model replaces several without loss, not that it is better.

## 5 Straight lines between seeds

Model soups work when two fine-tunes sit in one connected low-loss region, and a direct test is to walk along the line between them. Because both seeds start from the released weights and see the same rows, we expect them to be linearly connected. The test is cheap here: eleven points per pair, each evaluated on the full suite. Table 4 and Figure 4 give the result.

For v1 and its seed 1 the line shows a barrier: barrier +0.7 points [+0.1, +1.5]. For the name-randomised pair the general accuracy shows no barrier (the interval contains zero or lies below it) (+0.1 points [-0.6, +0.8]) and the specialist skill itself, swap-name accuracy, varies from 82.3% to 86.0% along the line (barrier +0.7 [-0.4,

+1.9]). In other words the skill is not a knife-edge solution that a different seed would miss: two independent trainings of the recipe are connected by a path on which the skill never drops below the lower endpoint by more than the noise.

The third panel adds a caution. Along the v1 line the model drifts from the left end to the right end gradually, and the midpoint disagrees with v1 on 4.2% and with seed 1 on 4.7% of decisions while the two endpoints disagree on 8.3%. The line does not pass through an unrelated function; it moves through functions that are in between.

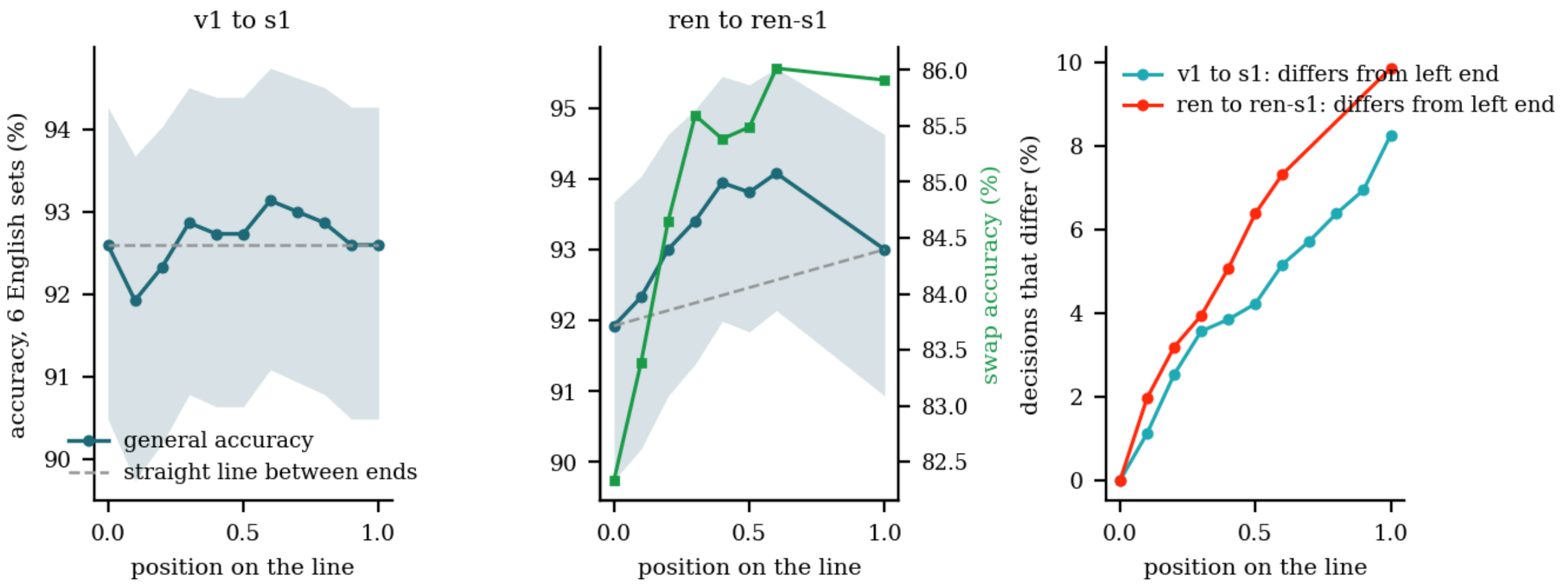

**Figure 4.** Straight lines between two seeds of the same recipe. Left and middle: general accuracy with Wilson bands, the dashed line joining the endpoints, and (middle, right axis) the swap-name accuracy of the name-randomised pair. Right: share of decisions that differ from the left endpoint along each line.

**Table 4.** Accuracy along the line between two seeds. Barrier = mean of the endpoint values minus the lowest interior value (positive: a dip); points are alpha = 0, 0.1, …, 1.

| Pair | Measure | Left end (%) | Right end (%) | Lowest interior (%) | Barrier, points [95% CI] | Points |
|---|---|---|---|---|---|---|
| v1 ↔ s1 | general accuracy | 92.6 | 92.6 | 91.9 (alpha 0.1) | +0.7 [+0.1, +1.5] | 11 |
| ren ↔ ren-s1 | general accuracy | 91.9 | 93.0 | 92.3 (alpha 0.1) | +0.1 [-0.6, +0.8] | 8 |
| ren ↔ ren-s1 | swap-name accuracy | 82.3 | 85.9 | 83.4 | +0.7 [-0.4, +1.9] | 8 |

## 6 Released weights, v1 and the specialists: lines

The line from the released weights to v1 is the one place where a merge is also a training trajectory's end points, so any nonlinearity shows how fast the fine-tune acts. Figure 5 and Table 5 give the curve. General accuracy rises from 72.9% at the released weights to 92.6% at v1. The path is concave: the halfway model is above the straight line by 6.9 points [5.1, 8.5], so most of the benefit of the fine-tune arrives in the first part of the move. There is no dip: no interior point is below the base. Extrapolating beyond v1 does not pay: the best extrapolated point (lambda 1.2) has 92.6%.

The multilingual set is a language shift inside the family: the training rows include 399 non-English rows, but the encoder behind the base is English-root. The base scores 61.2% on it, v1 73.6% and the halfway model 71.7%. Calibration moves with the weights: the ECE of the interior points lies between 2.9% and 31.2%, against 4.0% at v1, which is why we keep v1's stored temperature fixed for every merged model and report ECE as measured.

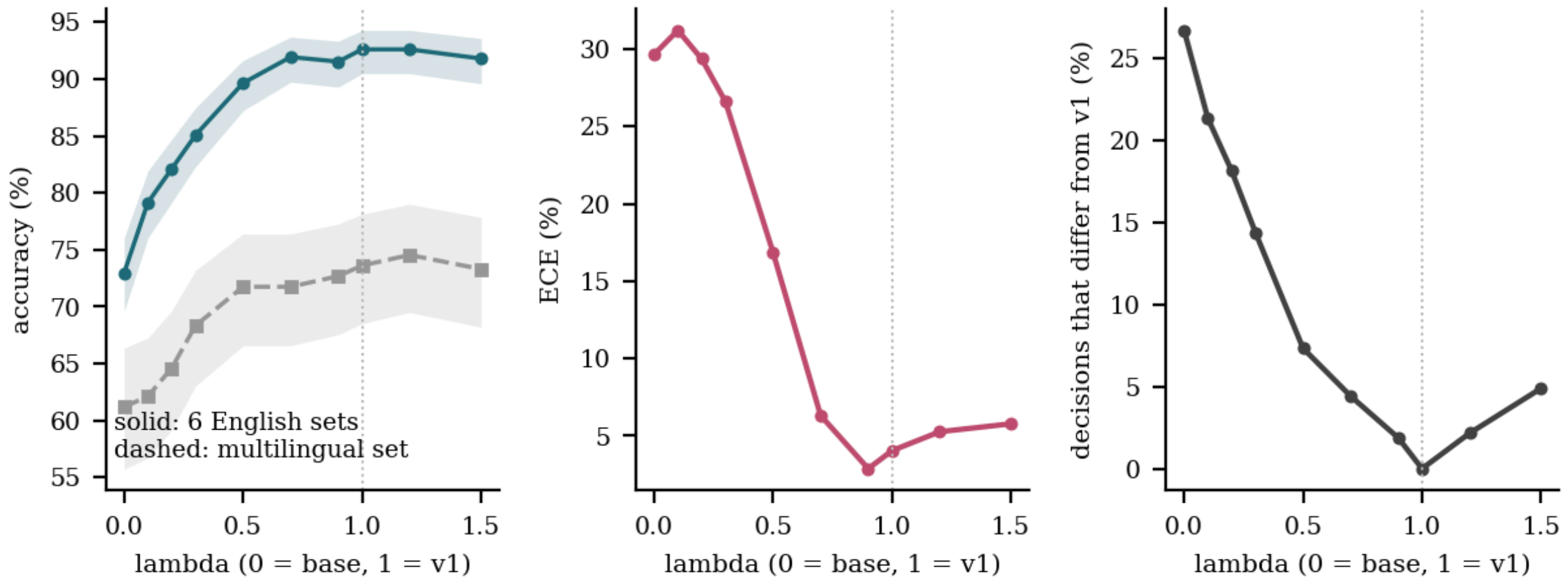

**Figure 5.** From the released weights (lambda = 0) to v1 (lambda = 1) and beyond. Left: accuracy on the six English sets (solid, Wilson band) and on the multilingual set (dashed). Middle: ECE. Right: decisions that differ from v1.

**Table 5.** Interpolation and extrapolation between the released weights and v1. "Above the line" is the accuracy minus the straight line between the two end accuracies, with a paired bootstrap interval.

| lambda | General accuracy (%) [95% CI] | Multilingual (%) | ECE (%) | Decisions differing from v1 (%) | Above the line (points) |
|---|---|---|---|---|---|
| 0 | 72.9 [69.6-76.0] | 61.2 | 29.6 | 26.7 | - |
| 0.1 | 79.1 [76.1-81.9] | 62.1 | 31.2 | 21.3 | +4.2 [+2.5, +5.9] |
| 0.2 | 82.1 [79.2-84.7] | 64.6 | 29.4 | 18.1 | +5.2 [+3.3, +7.1] |
| 0.3 | 85.1 [82.3-87.4] | 68.3 | 26.6 | 14.4 | +6.2 [+4.3, +8.2] |
| 0.5 | 89.6 [87.2-91.6] | 71.7 | 16.8 | 7.3 | +6.9 [+5.1, +8.5] |
| 0.7 | 91.9 [89.7-93.7] | 71.7 | 6.3 | 4.4 | +5.2 [+3.9, +6.5] |
| 0.9 | 91.5 [89.3-93.3] | 72.7 | 2.9 | 1.9 | +0.9 [-0.1, +1.8] |
| 1 | 92.6 [90.5-94.3] | 73.6 | 4.0 | 0.0 | - |
| 1.2 | 92.6 [90.5-94.3] | 74.5 | 5.3 | 2.2 | - |
| 1.5 | 91.8 [89.6-93.6] | 73.3 | 5.8 | 4.9 | - |

### 6.1 From the released weights to a specialist

Figure 6 repeats the experiment of the last section but measures each specialist's own skill and compares it with the plain v1 line. A specialist skill that is absent at the base can follow either of two shapes along the line: it can appear early and then plateau, or stay absent until the end. At lambda = 0.6 the share of the base-to-specialist skill gain already reached is: name-randomised 83%; HR-schema 82%. A value near 60% means a straight, proportional path; a value well above it means the skill arrives early, well below it means it arrives late.

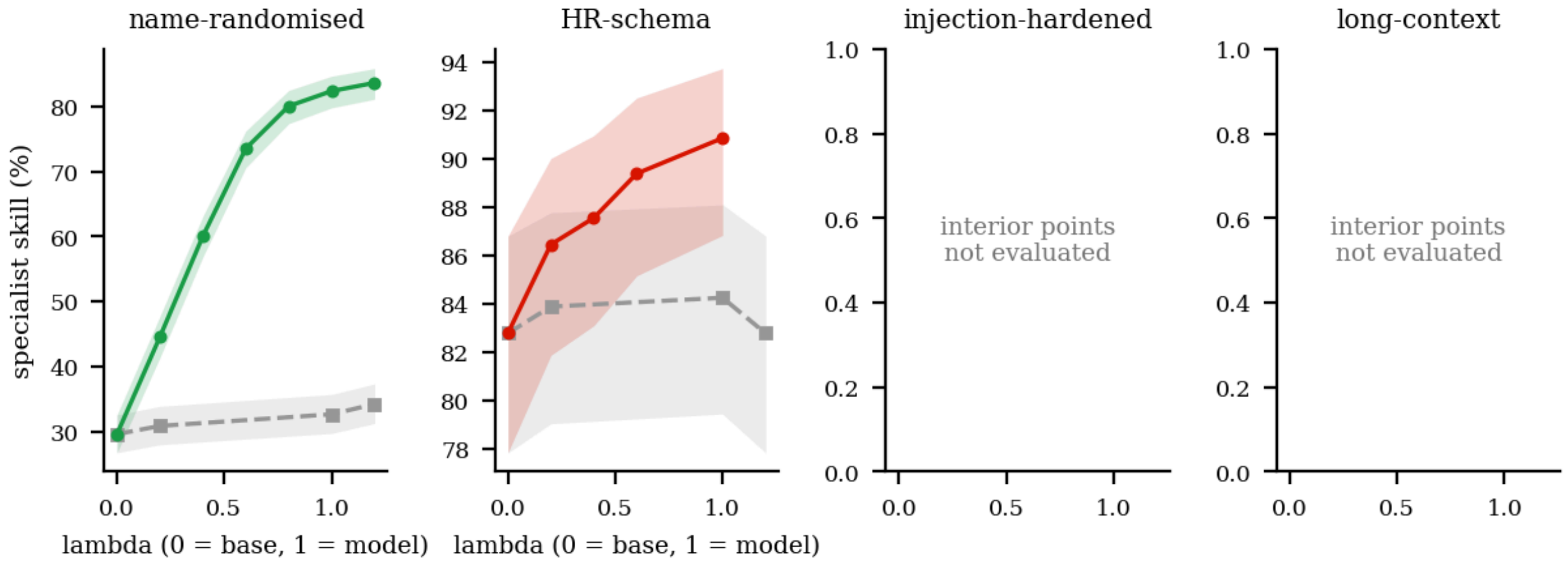

**Figure 6.** From the released weights to each specialist (solid, coloured) and to v1 (dashed grey), measured on the specialist's own skill, with Wilson bands. The no-shuffle variant has no interior points and is omitted.

### 6.2 From v1 to a specialist

The more useful line for a practitioner starts at the general model you already have. Figure 7 and Table 6 show what happens when v1 is moved a fraction $\alpha$ of the way to a specialist. At the halfway point (alpha = 0.5) the share of the specialist's gain that is kept is name-randomised 77% (general +0.9 points); HR-schema 83% (general +0.7 points); injection-hardened 33% (general +1.1 points); long-context 75% (general +0.1 points). A value of 50% would be a proportional path; larger values mean the skill is cheap to get, smaller values that it is concentrated in the last stretch. Pushing past the specialist ($\alpha$ = 1.5) gives gain-kept values of name-randomised 99%; HR-schema 72%; injection-hardened 111%; long-context 104% relative to the specialist, so extrapolation is sometimes useful. Moving the other way is the "negation" of a task vector: this was not evaluated.

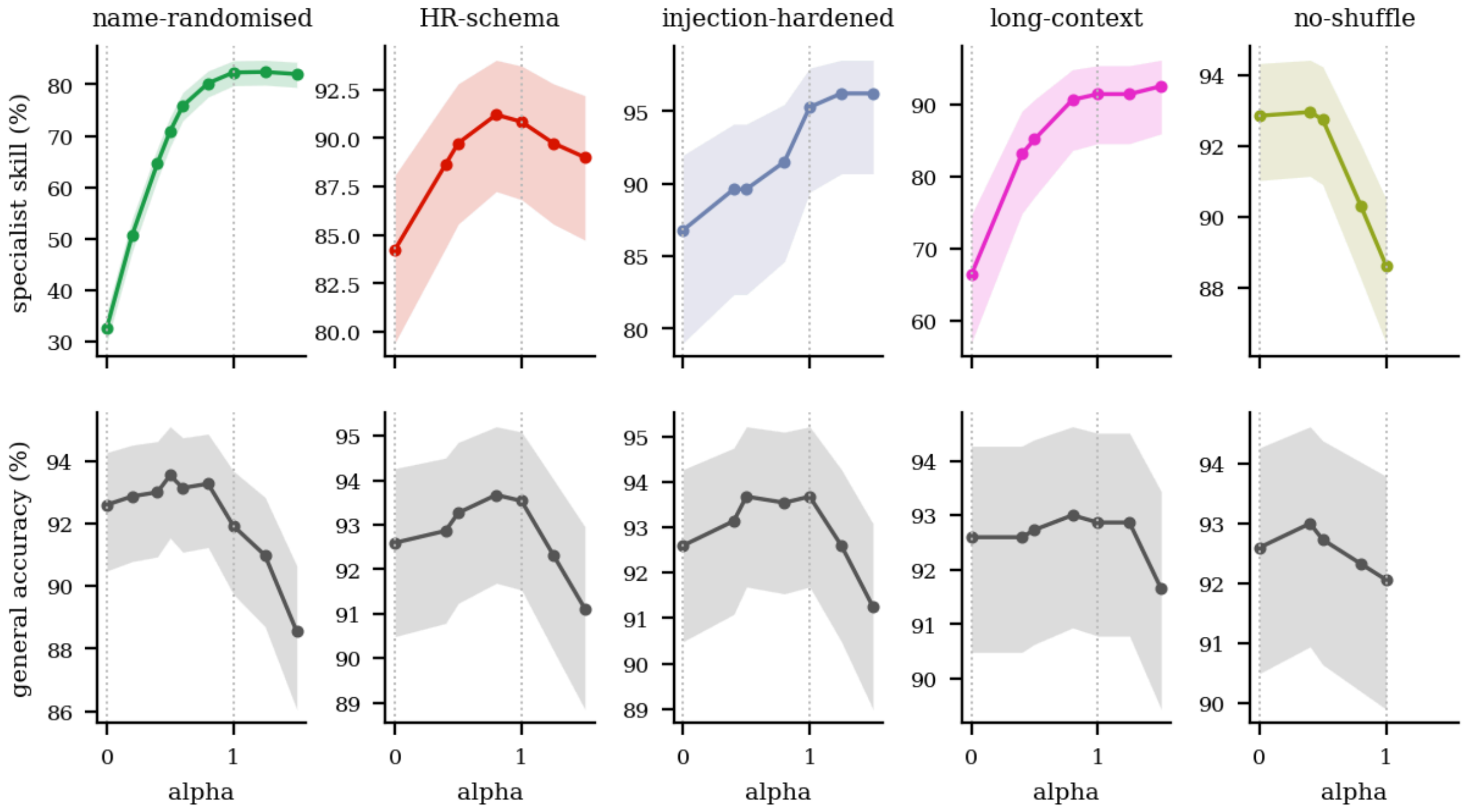

**Figure 7.** From v1 (alpha = 0) to each specialist (alpha = 1). Top: the specialist's own skill. Bottom: general accuracy on six English sets. Points to the left of 0 move away from the specialist, points to the right of 1 extrapolate past it. Wilson bands.

**Table 6.** Moving from v1 towards each specialist. Skill is the specialist's own measure; "gain kept" is the share of the specialist's gain over v1 [95% CI]; the last column is the change in general accuracy against v1 (paired bootstrap). For the no-shuffle variant the change in order stability is shown instead of a share.

| Specialist | alpha | Skill (%) [95% CI] | Gain kept (%) | General accuracy vs v1, points |
|---|---|---|---|---|

| Specialist | alpha | Skill (%) [95% CI] | Gain kept (%) | General accuracy vs v1, points |
|---|---|---|---|---|
| name-randomised | 0.5 | 70.9 [67.9-73.7] | 77 [72, 81] | +0.9 [-0.4, +2.3] |
| name-randomised | 1 | 82.3 [79.8-84.6] | 100 | -0.7 [-2.6, +1.2] |
| name-randomised | 1.5 | 82.0 [79.5-84.3] | 99 [96, 103] | -4.0 [-6.3, -1.7] |
| HR-schema | 0.5 | 89.7 [85.6-92.8] | 83 [55, 130] | +0.7 [-0.7, +2.0] |
| HR-schema | 1 | 90.8 [86.8-93.7] | 100 | +0.9 [-0.8, +2.7] |
| HR-schema | 1.5 | 89.0 [84.7-92.2] | 72 [13, 105] | -1.5 [-3.5, +0.5] |
| injection-hardened | 0.5 | 89.6 [82.4-94.1] | 33 [0, 70] | +1.1 [-0.1, +2.3] |
| injection-hardened | 1 | 95.3 [89.4-98.0] | 100 | +1.1 [-0.5, +3.1] |
| injection-hardened | 1.5 | 96.2 [90.7-98.5] | 111 [100, 150] | -1.3 [-3.5, +0.8] |
| long-context | 0.5 | 85.2 [77.2-90.7] | 75 [69, 82] | +0.1 [-1.3, +1.6] |
| long-context | 1 | 91.5 [84.6-95.5] | 100 | +0.3 [-1.6, +2.0] |
| long-context | 1.5 | 92.6 [86.0-96.2] | 104 [100, 109] | -0.9 [-3.0, +1.2] |
| no-shuffle | 0.5 | 92.7 [90.9-94.2] | -0.1 pts vs v1 | +0.1 [-1.2, +1.7] |
| no-shuffle | 1 | 88.6 [86.5-90.5] | -4.2 pts vs v1 | -0.5 [-2.4, +1.3] |

**Table 7.** Barrier along the line from v1 to each specialist for general accuracy and for the specialist skill (positive = interior dip below the endpoint mean).

| Specialist | Points on the line | General accuracy barrier [95% CI] | Skill barrier [95% CI] |
|---|---|---|---|
| name-randomised | 7 | -0.6 [-1.3, +0.5] | +6.8 [+4.4, +9.2] |
| HR-schema | 5 | +0.2 [-0.6, +1.1] | -1.1 [-3.1, +0.7] |
| injection-hardened | 5 | +0.0 [-0.7, +0.9] | +1.4 [-1.4, +3.8] |
| long-context | 5 | +0.1 [-0.5, +1.0] | -4.2 [-6.5, -2.2] |
| no-shuffle | 5 | -0.0 [-0.8, +0.9] | +0.4 [-0.9, +1.8] |

Of the 5 lines, 0 show a general-accuracy barrier whose interval excludes zero (none) and 1 show a barrier in the specialist skill itself (name-randomised). A line from v1 to a specialist is therefore not entirely free of interior dips; see the table for where.

## 7 Task arithmetic of the specialists

Task arithmetic adds specialist vectors to v1. With one specialist this is the line of Section 6 with $\alpha$=$\lambda$; with several, the specialists share the weights and may or may not get in each other's way. We evaluated every combination of the four specialists with a skill that can be measured as a gain (name-randomised, HR, injection, long-context) at $\lambda = 1/k$ and $\lambda = 1$, a sweep of eight values for the four-way sum, and a 5-way sum that includes the no-shuffle model. The no-shuffle model is left out of the headline numbers because its "skill" is lower than v1's.

The four-way average (lambda = 0.25) keeps 52 [38, 64]% of the gains on average; adding the four full vectors (lambda = 1) keeps 42 [3, 63]%, with a general accuracy change of -9.0 [-11.7, -6.2] points. Figure 8 shows how the retained gain and general accuracy move with $\lambda$, and Table 8 lists the values.

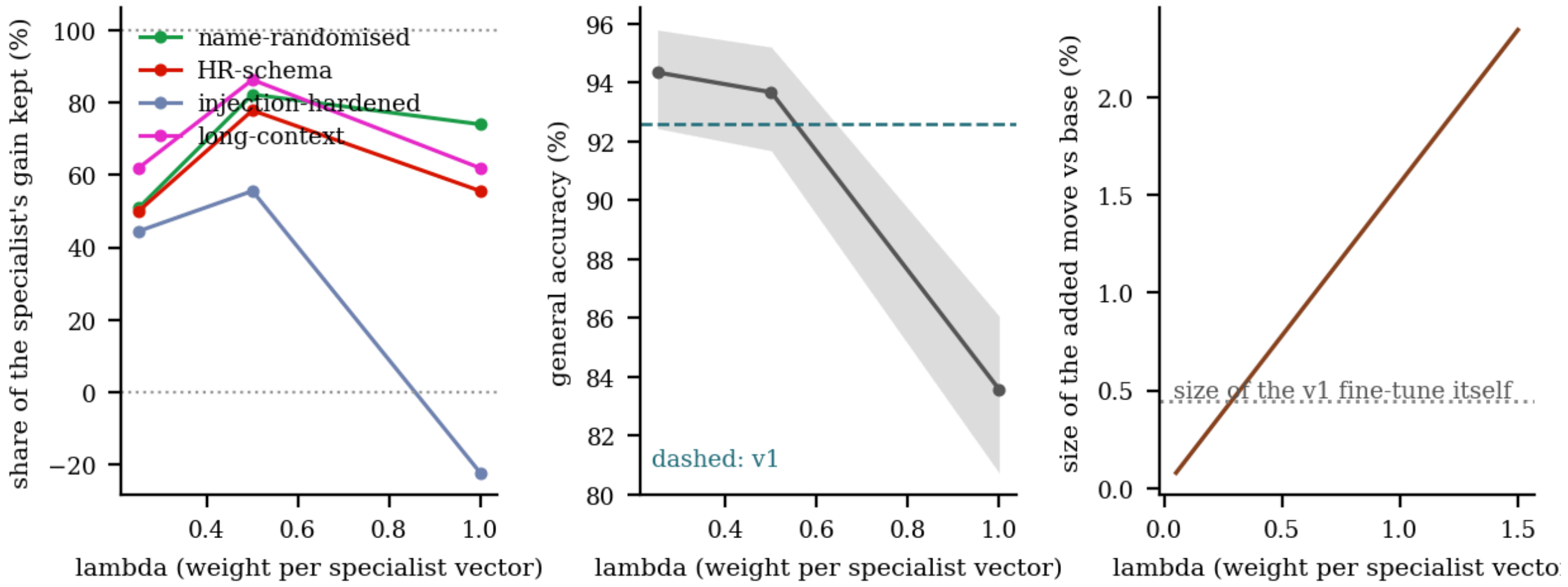

**Figure 8.** Adding the four specialist vectors to v1 with weight lambda each. Left: share of each specialist's gain that is kept. Middle: general accuracy (Wilson band; dashed: v1). Right: size of the added move relative to the size of the weights, against the size of v1's own fine-tune change.

**Table 8.** All four specialists added to v1, as a function of lambda (weight per vector). Gain kept per skill (%), their mean with a bootstrap interval, the change in general accuracy against v1 (points), ECE and decisions flipped against v1.

| lambda | rename | HR | inject | long | Mean kept (%) [95% CI] | General vs v1 [95% CI] | ECE (%) | Flipped (%) |
|---|---|---|---|---|---|---|---|---|
| 0.25 | 51 | 50 | 44 | 62 | 52 [38, 64] | +1.7 [+0.3, +3.4] | 2.6 | 8.8 |
| 0.5 | 82 | 78 | 56 | 86 | 75 [62, 88] | +1.1 [-0.8, +3.0] | 3.8 | 13.4 |
| 1 | 74 | 56 | -22 | 62 | 42 [3, 63] | -9.0 [-11.7, -6.2] | 10.3 | 24.1 |

Figure 9 and Table 9 break the result down by combination. The first panel is the safe setting, the average of the specialists. The retained gain on a skill falls as more specialists share the same total weight, which is the expected cost of averaging: each vector gets only $1/k$ of its length. The second panel, full vectors, tells us how far the vectors really are from interfering: if retention at $\lambda$=1 is close to 100% for the skills contained and close to 0% elsewhere, the vectors can be added with little interference; deviations from that are interference or spill-over.

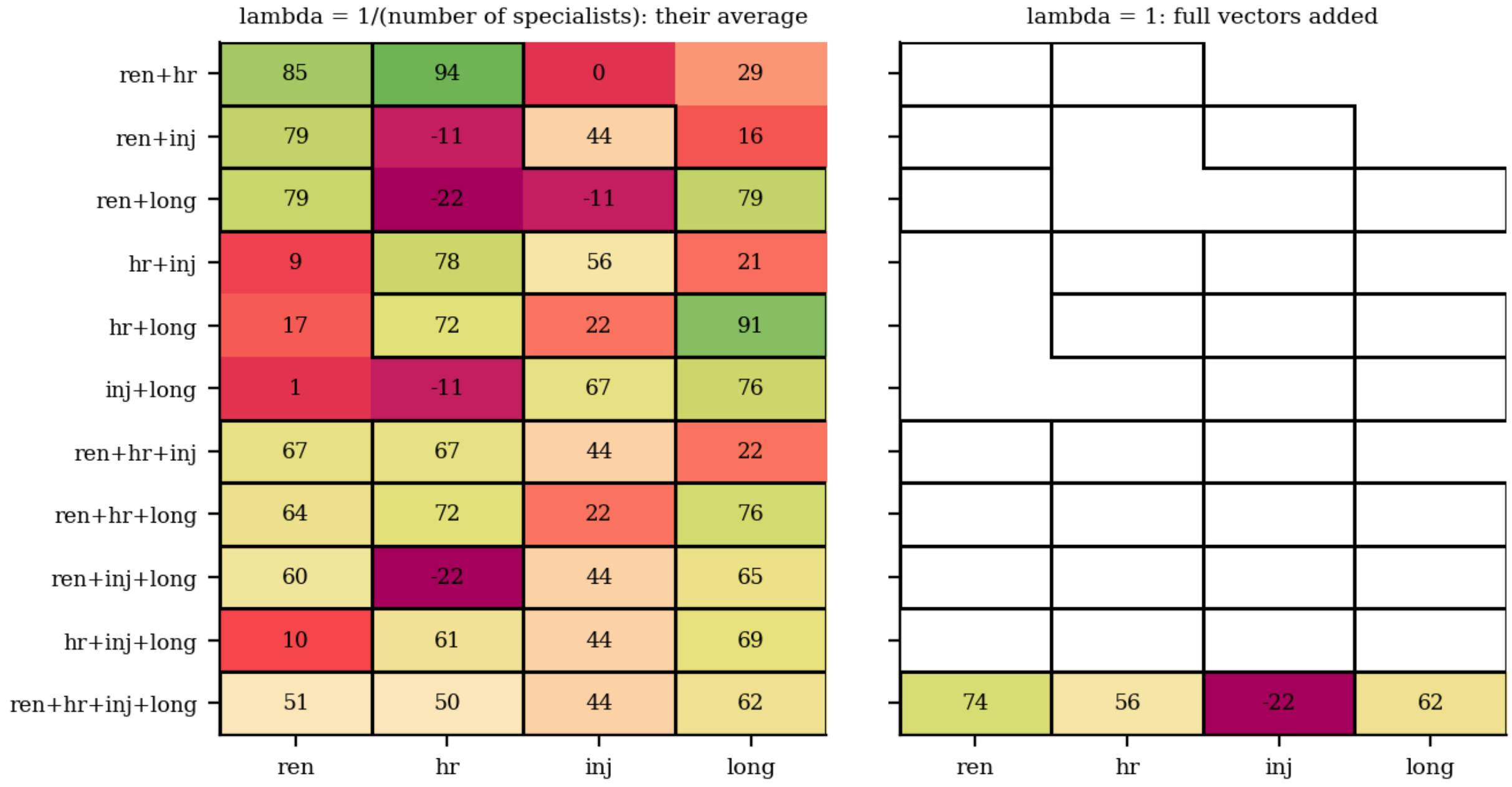

**Figure 9.** Share of each specialist's gain kept (%) by every combination of two, three and four specialists added to v1. Left: lambda = 1/k (their average). Right: lambda = 1 (full vectors). Outlined cells are the skills the combination contains; the others show spill-over.

**Table 9.** Combinations of k specialists, by weight. "Own" is the mean gain kept on the skills the combination contains (mean, then range over combinations); "spill-over" is the mean on the skills it does not contain; the last column is the change in general accuracy against v1 in points.

| k | Weight | Combinations | Own skills kept (%) | Spill-over (%) | General vs v1, points |
|---|---|---|---|---|---|
| 2 | average | 6 | 75 (62 to 90) | 5 | +1.2 (+0.4 to +1.5) |
| 3 | average | 4 | 61 (56 to 70) | 8 | +1.7 (+1.2 to +2.4) |
| 4 | average | 1 | 52 (52 to 52) | - | +1.7 (+1.7 to +1.7) |
| 4 | full | 1 | 42 (42 to 42) | - | -9.0 (-9.0 to -9.0) |

Averaged over all 12 ordered pairs, adding a second specialist at half weight changes a skill by +1.4 points relative to the single-specialist line at the same alpha; the interval excludes zero for 2 of 12. Table 10 lists the twelve ordered pairs.

**Table 10.** Effect of a partner on a specialist skill: skill of the two-specialist average (lambda = 0.5 each) against the same skill on the v1-to-specialist line at alpha = 0.5.

| Skill | Partner | Change, points [95% CI] |
|---|---|---|
| name-randomised | HR-schema | +4.2 [+2.5, +5.9] |
| name-randomised | injection-hardened | +1.2 [-0.5, +2.8] |
| name-randomised | long-context | +0.9 [-0.9, +2.8] |
| HR-schema | name-randomised | +0.7 [-0.7, +2.2] |
| HR-schema | injection-hardened | -0.4 [-1.1, +0.0] |
| HR-schema | long-context | -0.7 [-2.2, +0.4] |
| injection-hardened | name-randomised | +0.9 [-1.9, +3.8] |
| injection-hardened | HR-schema | +1.9 [+0.0, +4.7] |
| injection-hardened | long-context | +2.8 [+0.0, +6.6] |
| long-context | name-randomised | +1.1 [-0.6, +3.1] |
| long-context | HR-schema | +4.1 [+2.7, +5.8] |
| long-context | injection-hardened | +0.3 [-0.6, +1.3] |

### 7.1 Do the vectors transfer to other seeds?

A specialist vector was computed against one particular v1 run. If it is the same change in every run, adding it to another v1 seed should install the skill. If it was entangled with that run's noise, it should not. Figure 10 and Table 11 show the result for seeds 1 and 2.

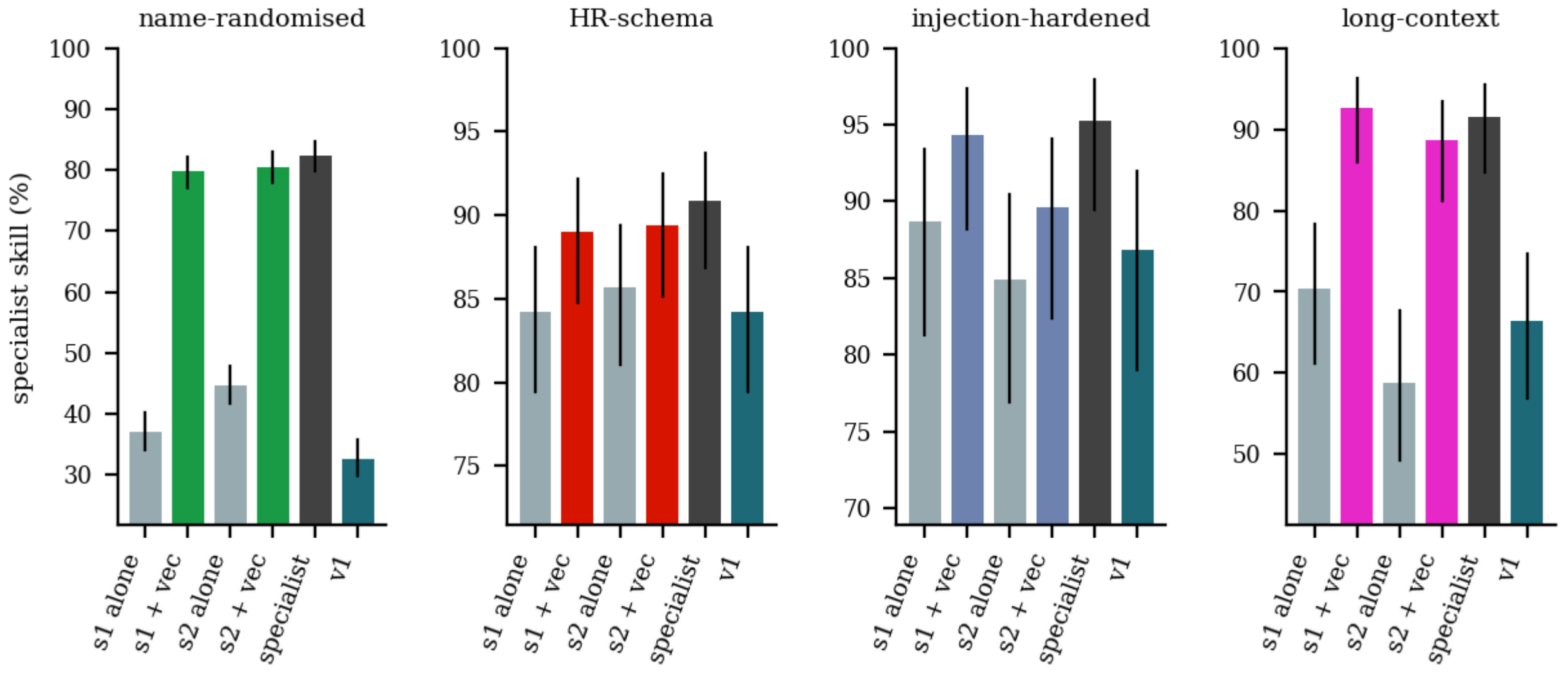

**Figure 10.** Is a specialist vector portable? The vector (specialist minus v1) is added to seeds 1 and 2 of the plain recipe, which never saw the specialist rows. Bars: skill of the seed alone, of the seed plus the vector, of the specialist and of v1, with Wilson intervals.

**Table 11.** Adding a specialist vector (specialist minus v1) to another plain seed. "Gain" is the skill increase over the seed as a share of the specialist's increase over v1; the last column is the change in general accuracy against the seed alone.

| Specialist | Seed | Seed alone: skill (%) | Seed + vector: skill (%) | Specialist: skill (%) | Gain (%) | General vs seed, points |
|---|---|---|---|---|---|---|
| name-randomised | s1 | 37.1 [34.1-40.2] | 79.8 [77.1-82.2] | 82.3 [79.8-84.6] | 86 | -1.6 [-3.4, +0.3] |
| name-randomised | s2 | 44.8 [41.7-48.0] | 80.5 [77.9-82.9] | 82.3 [79.8-84.6] | 72 | -3.9 [-5.8, -2.2] |
| HR-schema | s1 | 84.2 [79.5-88.1] | 89.0 [84.7-92.2] | 90.8 [86.8-93.7] | 72 | -1.3 [-3.0, +0.3] |
| HR-schema | s2 | 85.7 [81.1-89.4] | 89.4 [85.2-92.5] | 90.8 [86.8-93.7] | 56 | -1.6 [-3.4, +0.3] |
| injection-hardened | s1 | 88.7 [81.2-93.4] | 94.3 [88.2-97.4] | 95.3 [89.4-98.0] | 67 | -0.1 [-1.9, +1.5] |
| injection-hardened | s2 | 84.9 [76.9-90.5] | 89.6 [82.4-94.1] | 95.3 [89.4-98.0] | 56 | -1.2 [-2.8, +0.5] |
| long-context | s1 | 70.4 [61.2-78.3] | 92.6 [86.0-96.2] | 91.5 [84.6-95.5] | 88 | -1.1 [-2.7, +0.7] |
| long-context | s2 | 58.8 [49.3-67.7] | 88.7 [81.2-93.4] | 91.5 [84.6-95.5] | 119 | -3.1 [-4.8, -1.3] |

Averaged over the two seeds, the vector installs 79% of the name-randomised gain, 64% of the HR-schema gain, 61% of the injection-hardened gain, 103% of the long-context gain. The long-context vector transfers fully; the other three transfer partly, so the vectors are mostly but not fully portable between runs. Section 14 adds the other direction: replacing the rename vector by the one from another seed changes the four-way merge.

## 8 Interference: size, direction and sign conflicts

The recipes above work or fail because of the geometry of the changes they make to the weights. We computed that geometry exactly from the stored weights: all inner products between the 13 task vectors, by encoder layer, and sign statistics for every pair of specialists. Distances are in Appendix B.

### 8.1 Size

The v1 fine-tune moves the weights by 0.44% of their own norm. A specialist differs from v1 by 0.46-0.52% (Table 12), against 0.47% between two v1 seeds. The specialist-specific change is therefore 1.0-1.1 times the size of seed noise, and 0.9 times smaller than the general fine-tune. A vector that is mostly seed noise would be useless to add to another model; one that is much larger than seed noise is likelier to be a real change.

Figure 13 shows that the change is not spread evenly: name-randomised peaks at layer 8 (1.2 times the seed difference there) and exceeds it in 28 of 28 layers; HR-schema peaks at layer 14 (1.1 times the seed difference there) and exceeds it in 28 of 28 layers; injection-hardened peaks at layer 10 (1.1 times the seed difference there) and exceeds it in 28 of 28 layers; long-context peaks at layer 23 (1.1 times the seed difference there) and exceeds it in 28 of 28 layers; no-shuffle peaks at layer 1 (1.0 times the seed difference there) and exceeds it in 0 of 28 layers. The head, scorer and act head (6.3% of the parameters) hold 49-52% of the squared change and the embedding table 1-1%.

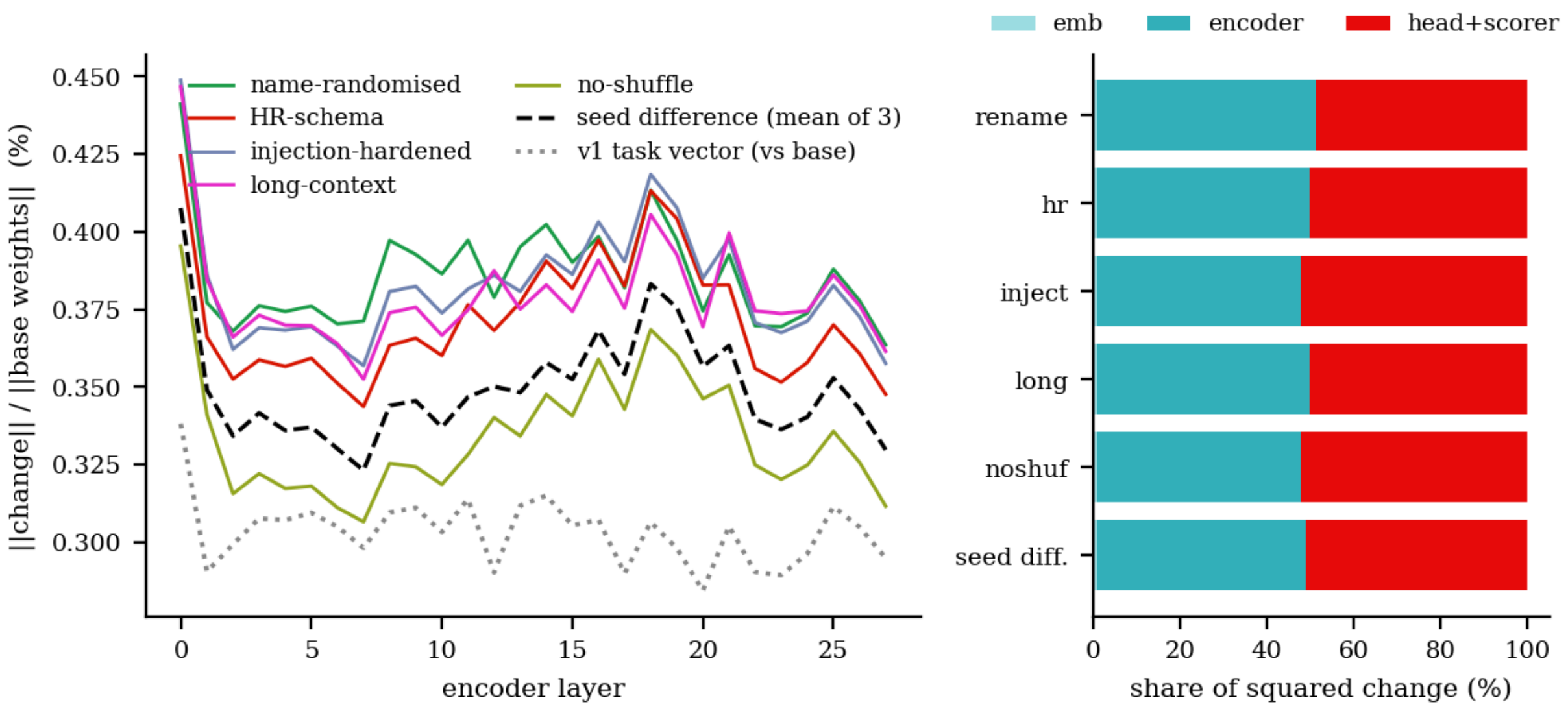

**Figure 13.** Size of each change relative to the size of the weights, by encoder layer (left), and where the squared change sits (right). The dashed line is the difference between two seeds of the same recipe; any specialist change near or below it is not separable from training noise by size alone.

**Table 12.** Size of each weight change relative to the norm of the released weights, and relative to the typical difference between two seeds of the same recipe.

| Change | Norm / norm of weights (%) | Ratio to a seed difference |
|---|---|---|
| name-randomised | 0.51 | 1.08 |
| HR-schema | 0.50 | 1.06 |
| injection-hardened | 0.52 | 1.11 |
| long-context | 0.51 | 1.08 |
| no-shuffle | 0.46 | 0.97 |
| seed difference (mean of 3 seed pairs) | 0.47 | 1.00 |
| v1 fine-tune itself (vs released weights) | 0.44 | 0.9 |

## 8.2 Direction

The task vectors of all fine-tunes against the released weights are strongly aligned: the mean off-diagonal cosine is 0.41 (Figure 11), and seeds of one recipe are hardly closer to each other (0.42) than to models of other recipes (0.41). Every fine-tune on these rows makes much the same move, and most of it is not specific to a recipe. The specialist-specific part has to be read against a reference. All specialist vectors are measured against the same v1, so they share its seed noise: two *plain* seeds, each minus v1, have a cosine of 0.50 although they have nothing in common but that reference. Between different specialists the cosine is 0.45 on average (range 0.42 to 0.47, Figure 12), no higher than between two plain seeds. Cosine by itself therefore cannot tell a specialist vector from a seed difference.

A cleaner test uses the specialists that have more than one seed. The change made by the name-randomised recipe, measured each time against the plain seed with the same number, has a cosine of 0.04 between seeds, and the no-shuffle recipe 0.02, against 0.02 for a control that pairs a vector of one recipe with a vector of the other (and a different seed). That is a small but repeatable direction: if a vector is a signal plus independent noise, its cosine across seeds estimates the share of its squared length that is signal, so only about 3% of a specialist vector is the recipe and the rest is seed noise. The skill is in that few per cent, and a merge that keeps it has to keep it against a much larger noise.

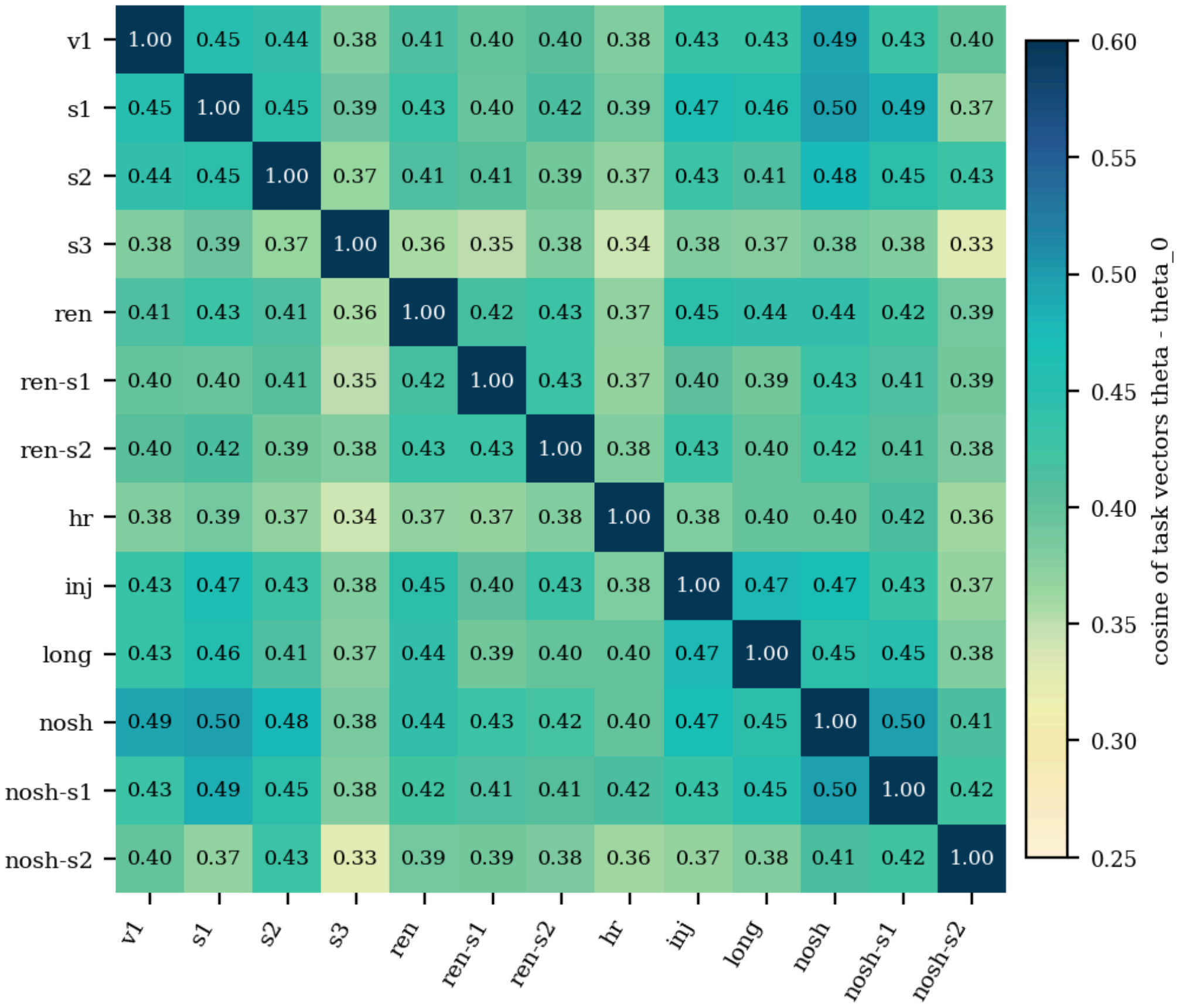

**Figure 11.** Cosine between the task vectors (fine-tune minus released weights) of every pair of fine-tunes, over all parameters. Seeds of one recipe are barely closer to each other than to other recipes: most of the shared movement is what any fine-tune on these rows does.

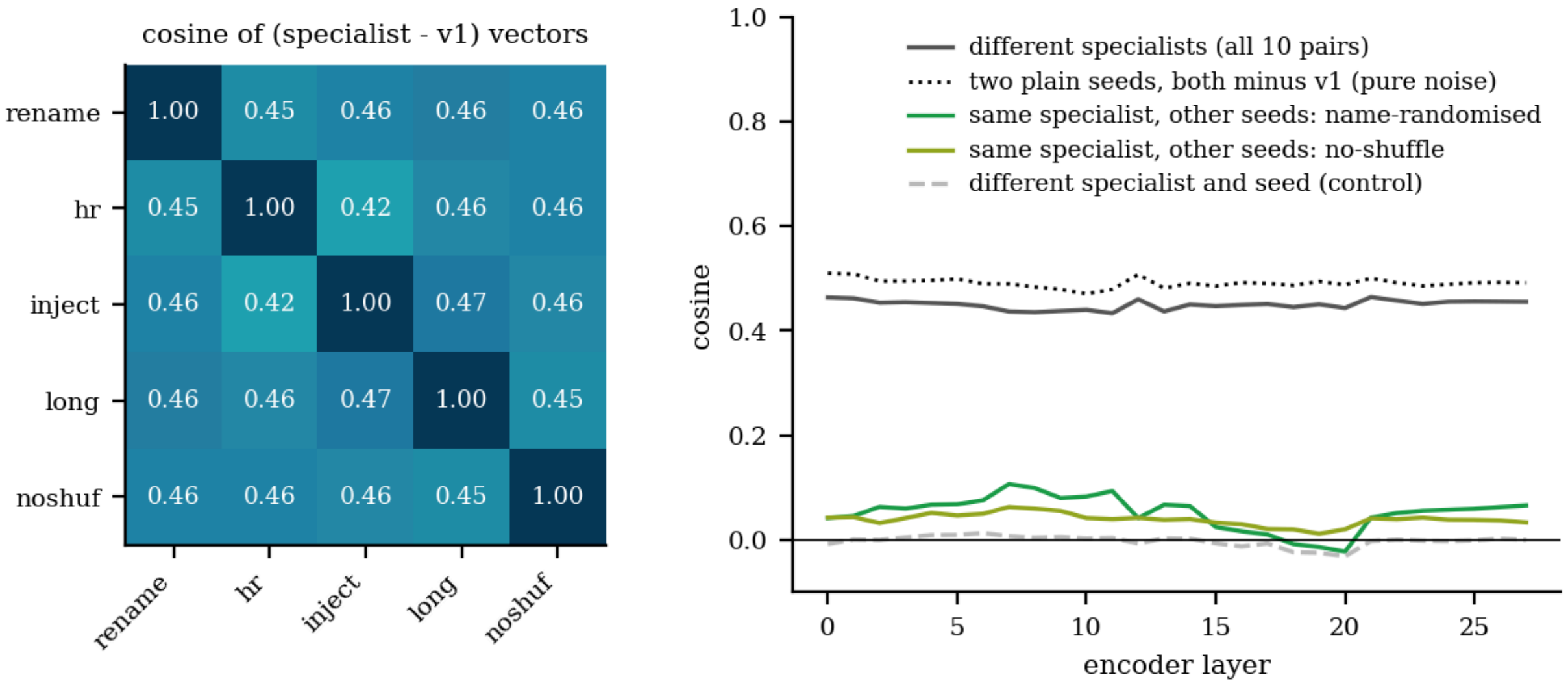

**Figure 12.** Directions of the specialist-specific change (specialist minus v1). Left: cosine between the five specialists. Right: cosine by encoder layer between different specialists, and between the same specialist trained with a different seed (a control for how much of a vector is seed noise).

### 8.3 Sign conflicts

TIES was proposed because averaging fails when different tasks push the same weight in opposite directions. For each pair of specialists we counted the coordinates that both changed (337-341 million of 421 million, because in half precision many small updates round to nothing) and the fraction of them where the signs agree. Over all such coordinates the agreement is 66.5% on average (range 65.8% to 67.8%) and over the coordinates in the top 20% by magnitude for both it is 88.4% (87.3% to 90.0%; Figure 14). Both are far above the 50% of chance, which would look like cooperation between the vectors. It is not: for two Gaussian vectors with cosine $c$ the expected sign agreement is $1/2 + \arcsin(c)/\pi$, and with each pair's own cosine that expectation is 65.0% on average, within 2.6 points of what was observed. The agreement is what a shared v1 reference produces, not evidence that specialists reinforce each other. About 12% of the large coordinates do conflict in sign, which is the amount TIES has to resolve.

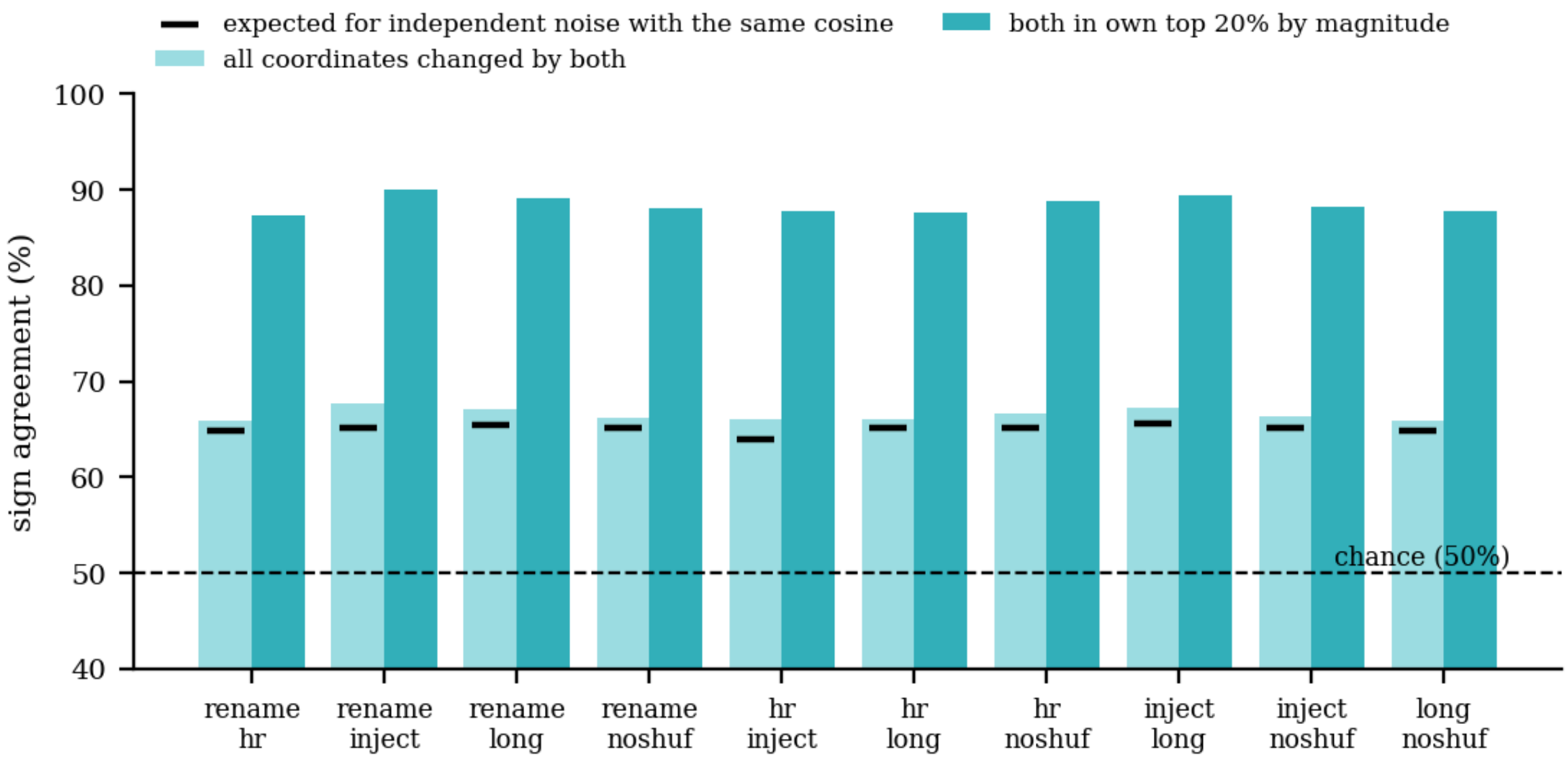

**Figure 14.** Sign agreement between the specialist-specific changes (specialist minus v1) of each pair of specialists, over all coordinates that both changed and over the coordinates in the top 20% by magnitude for both. Black ticks: what independent noise sharing only the v1 reference would give at the pair's own cosine. Chance is 50%.

## 9 TIES-style merging

TIES-style merging changes how the vectors are combined: trim each to its largest entries, elect a sign, and average only the entries that agree with it. We implemented it per weight tensor on the four specialist vectors (specialist minus root) and ran a reduced grid: rooted at v1, keeping 20% of each vector at scales 0.5, 1 and 1.5 and keeping 50% at scale 1; rooted at the released weights, keeping 20% at scale 1; plus two ablations rooted at v1, trimming without sign election and sign election without trimming. The disjoint mean divides by the number of vectors that agree at a coordinate, so TIES at $\lambda$=1 is much larger than a plain average and is comparable to a sum of the vectors where they do not overlap. The best TIES setting rooted at v1 is "TIES from v1, keep 20%, lambda 1.5" with a mean gain kept of 76% [63, 89] and a general-accuracy change of +1.1 [-0.8, +2.8] points. For comparison the plain average of the four vectors keeps 52% [38, 64] and the plain sum of full vectors keeps 42% [3, 63].

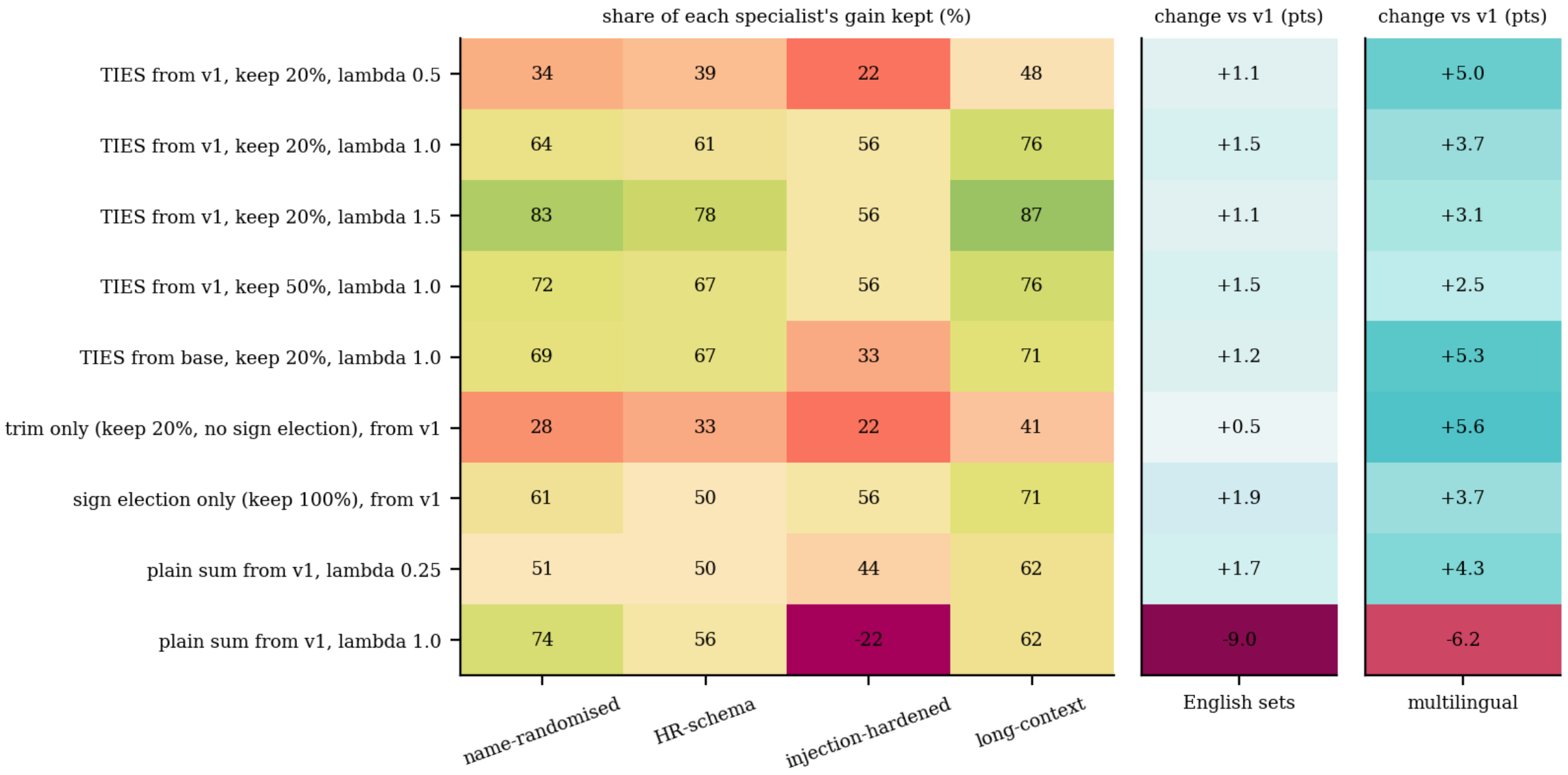

**Figure 15.** TIES-style merges of the four specialists. Left: share of each specialist's gain kept (%), by trim density, scale lambda and root. Right: change in general and multilingual accuracy against v1 (points). Last rows: plain sums for reference.

**Table 13.** TIES-style merges of the four specialists and plain sums, with bootstrap intervals on the mean gain kept.

| Merge | Mean gain kept (%) [95% CI] | General vs v1 [95% CI] | ECE (%) | Flipped (%) |
|---|---|---|---|---|
| TIES from v1, keep 20%, lambda 0.5 | 36 [26, 49] | +1.1 [-0.3, +2.6] | 2.8 | 5.5 |
| TIES from v1, keep 20%, lambda 1.0 | 64 [50, 76] | +1.5 [-0.3, +3.2] | 3.2 | 11.0 |
| TIES from v1, keep 20%, lambda 1.5 | 76 [63, 89] | +1.1 [-0.8, +2.8] | 4.0 | 13.4 |
| TIES from v1, keep 50%, lambda 1.0 | 68 [53, 81] | +1.5 [-0.3, +3.2] | 3.4 | 11.5 |
| TIES from base, keep 20%, lambda 1.0 | 60 [44, 76] | +1.2 [-0.5, +3.1] | 4.3 | 10.9 |
| trim only (keep 20%, no sign election), from v1 | 31 [21, 44] | +0.5 [-0.7, +1.7] | 3.0 | 5.2 |
| sign election only (keep 100%), from v1 | 59 [44, 72] | +1.9 [+0.1, +3.6] | 2.7 | 10.4 |
| plain sum from v1, lambda 0.25 | 52 [38, 64] | +1.7 [+0.3, +3.4] | 2.6 | 8.8 |
| plain sum from v1, lambda 1.0 | 42 [3, 63] | -9.0 [-11.7, -6.2] | 10.3 | 24.1 |

Trimming alone (20% kept, no election) keeps 31% of the gains on average and changes general accuracy by +0.5 points. Sign election alone (no trimming) keeps 59% and changes general accuracy by +1.9 points. Rooted at the released weights, the TIES merge carries the general fine-tune with it: general accuracy for that row is 93.8% (v1: 92.6%).

## 10 Which layers carry the specialist change

If a specialist skill lives in a narrow part of the network, we could copy only that part and leave the rest as v1; this would also avoid interference with other specialists that live elsewhere. We tested four masks per specialist in a reduced plan: the head, scorer and act head together; the top seven layers with the final norm; the lower and upper halves of the encoder (the lower half with the embeddings, the upper half with the final norm). The lower seven layers alone and the all-attention and all-MLP masks were not run. In every case the masked parameters take the specialist's values and all the others stay at v1. Figure 16 and Table 14 show the gain that survives, next to the share of the weight change that each part holds.

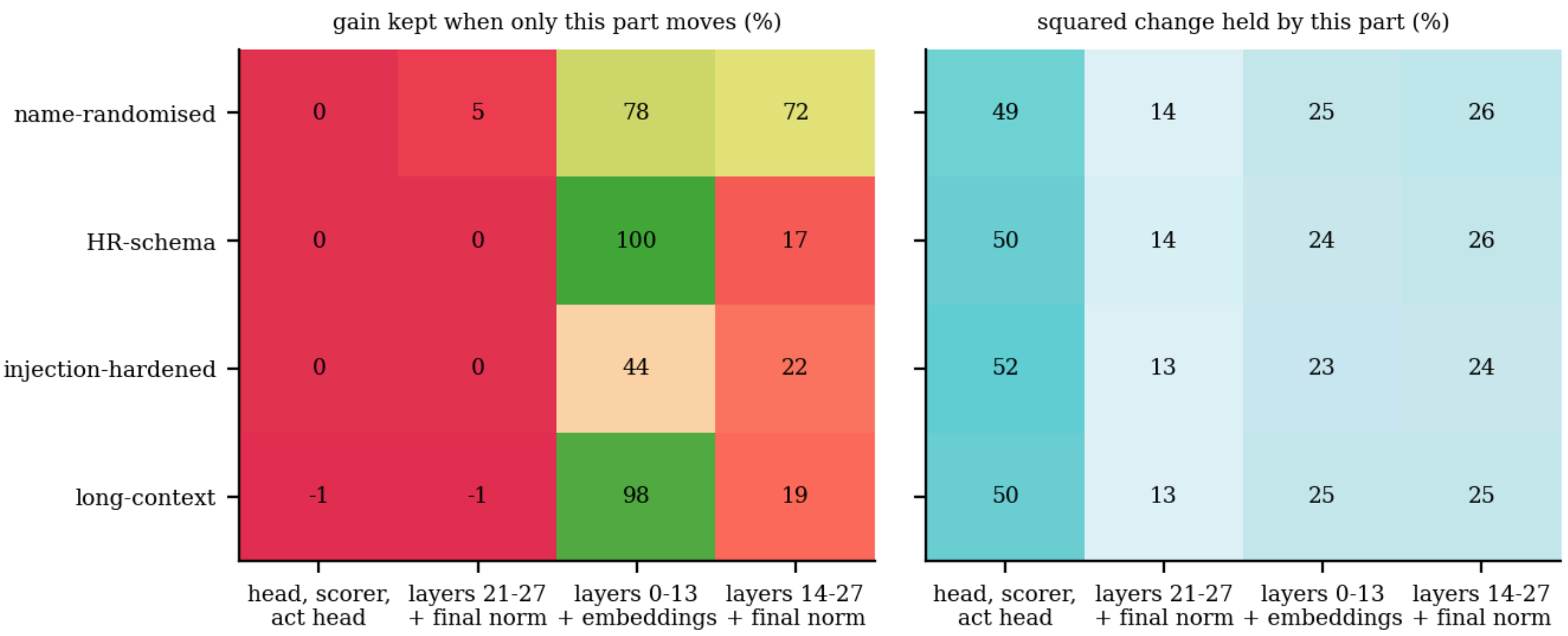

**Figure 16.** Which part of the network carries the specialist change? Left: share of each specialist's gain kept when only the parameters in the named part take the specialist's values and everything else stays at v1. Right: the share of the specialist's squared weight change that sits in that part.

**Table 14.** Gain kept (%) [95% CI] when only one part of the network is moved from v1 to the specialist.

| Specialist | head, scorer, act head | layers 21-27 + final norm | layers 0-13 + embeddings | layers 14-27 + final norm |
|---|---|---|---|---|
| name-randomised | 0 [-0, 1] | 5 [3, 7] | 78 [73, 81] | 72 [67, 77] |
| HR-schema | 0 [0, 0] | 0 [0, 0] | 100 [76, 145] | 17 [-27, 47] |
| injection-hardened | 0 [0, 0] | 0 [0, 0] | 44 [11, 80] | 22 [0, 60] |
| long-context | -1 [-2, 0] | -1 [-3, 2] | 98 [94, 102] | 19 [13, 27] |

The part that keeps the most gain, per specialist, is: name-randomised: layers 0-13 + embeddings (78%); HR-schema: layers 0-13 + embeddings (100%); injection-hardened: layers 0-13 + embeddings (44%); long-context: layers 0-13 + embeddings (98%). Moving only the head, scorer and act head (6.3% of the parameters) keeps 0%, 0%, 0%, -1% of the gain for the four specialists in the order of the table. A part that holds a large share of the change but keeps little of the gain is carrying noise; a part that holds a small share and keeps most of the gain is where the specialist change is concentrated. The comparison of the two panels is the answer to which layers matter.

## 11 When does merging work? A recipe table

Figure 17 puts every merged model that contains a specialist on one plane. A useful merge is in the upper right, with the share of the target gains kept near 100% and general accuracy not below v1. Table 15 condenses the experiments into a recipe table: the operation, what it keeps, what it costs, and a verdict from a rule fixed before the results were read (stated in the caption). Where a row averages several experiments the interval is the mean of the per-experiment intervals and so is only a rough guide; where a row picks the best setting from a sweep on the same test items, the number is an upper bound on what a fresh set would give.

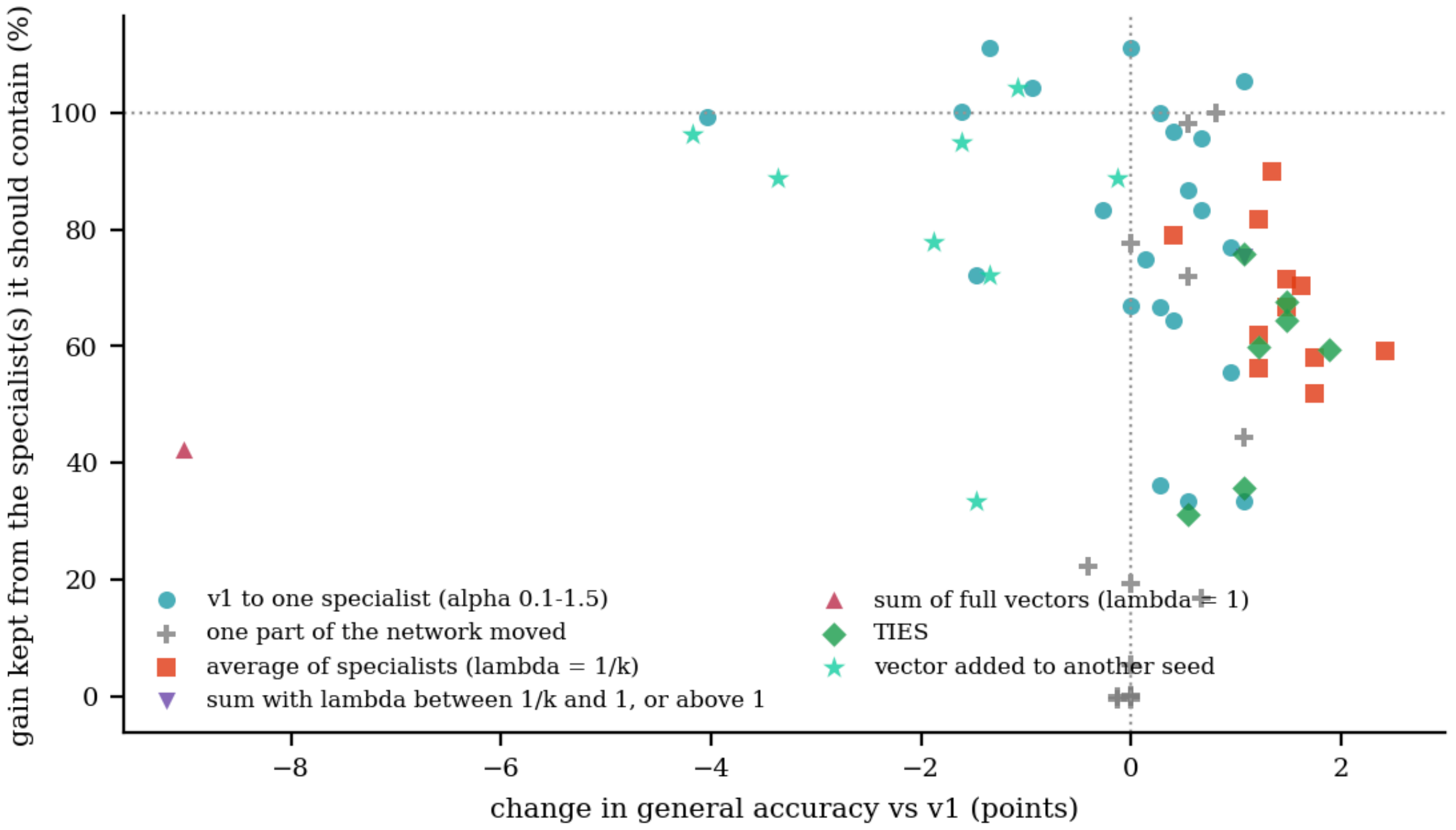

**Figure 17.** Every merged model that contains at least one specialist change: change in general accuracy against v1 (x) against the share of the target gains kept (y). The ideal merge sits at the top, at or right of zero.

**Table 15.** When does merging work? Mean gain kept over the targeted skills [95% CI], change in general accuracy against v1 (points [95% CI]), decisions flipped against v1 and ECE. Verdict rule, fixed in advance: works if the lower bound of the gain kept is at least 80% and the general accuracy interval stays above −2 points; partly works if the mean is at least 50% and general accuracy stays above −2 points; fails otherwise. Rows picked on the test items are optimistic.

| Recipe | Gain kept (%) | General vs v1 (points) | Flipped (%) | ECE (%) | Verdict |
|---|---|---|---|---|---|
| Move v1 halfway to one specialist (mean of the four) | 67 [49, 91] | +0.7 [-0.6, +2.1] | 5.3 | 3.3 | partly works |
| Average of 2 specialists, lambda 1/2 (mean over 6 combinations) | 75 [60, 93] | +1.2 [-0.4, +2.9] | 9.6 | 3.2 | partly works |
| Average of 3 specialists, lambda 1/3 (mean over 4 combinations) | 61 [47, 77] | +1.7 [+0.2, +3.4] | 9.0 | 2.7 | partly works |
| Average of all four specialists, lambda 1/4 | 52 [38, 64] | +1.7 [+0.3, +3.4] | 8.8 | 2.6 | partly works |
| Sum of all four full vectors, lambda 1 | 42 [3, 63] | -9.0 [-11.7, -6.2] | 24.1 | 10.3 | fails (general accuracy drops) |
| Best four-way lambda with no general loss (lambda 0.5; picked on the test items) | 75 [62, 88] | +1.1 [-0.8, +3.0] | 13.4 | 3.8 | partly works |
| Best TIES setting from v1 (TIES from v1, keep 20%, lambda 1.5; picked on the test items) | 76 [63, 89] | +1.1 [-0.8, +2.8] | 13.4 | 4.0 | partly works |
| Add one specialist vector to another plain seed (mean over 8 cases) | 77 [77, 77] | -1.7 [-3.5, -0.0] | - | - | fails (general accuracy drops) |
| Move only the head, scorer and act head to the specialist (mean over the four) | -0 [-1, 0] | +0.0 [+0.0, +0.0] | 0.0 | 4.1 | fails (skills not kept) |

By the rule, 0 of 9 recipes work; 6 work partly: Move v1 halfway to one specialist (mean of the four); Average of 2 specialists, lambda 1/2 (mean over 6 combinations); Average of 3 specialists, lambda 1/3 (mean over 4 combinations); Average of all four specialists, lambda 1/4; Best four-way lambda with no general loss (lambda 0.5; picked on the test items); Best TIES setting from v1 (TIES from v1, keep 20%, lambda 1.5; picked on the test items); 3 fail: Sum of all four full vectors, lambda 1; Add one specialist vector to another plain seed (mean over 8 cases); Move only the head, scorer and act head to the specialist (mean over the four).

## 12 Joint training and prediction ensembling as the reference

The comparison that a merge should be judged against is a single fine-tune on the union of the data. We trained it: the joint model starts from the released weights, uses the same trainer and hyper-parameters as every other model (four epochs, options shuffled, same learning rates) on the 5,544 rows made of the v1 rows and the extra rows of the four specialists, and was evaluated on exactly the suites of Section 3. It was trained with two seeds, so the table shows how much joint training itself varies. We also add the baseline that the weight-averaging literature uses to explain a merge, an ensemble of the same models that averages their output probabilities instead of their weights. The ensemble needs one forward pass per member at test time, so it is not a replacement for a merge, but it separates two questions: whether the information is in the five models, and whether weight space can hold it in one.

With seed 0, joint fine-tuning keeps 120% [104, 162] of the four specialist gains, with a general accuracy change of +0.3 [-1.3, +2.2] points. With seed 1, joint fine-tuning keeps 116% [104, 149] of the four specialist gains, with a general accuracy change of +0.9 [-0.7, +2.7] points. A joint model can keep more than 100% because it sees every specialist's extra rows, while each specialist sees only its own. By skill, for seed 0: name-randomised 101% (average 51%); HR-schema 139% (average 50%); injection-hardened 122% (average 44%); long-context 118% (average 62%). Against the four-way weight average its mean retention is +68 points [+51, +108] and against the sum of full vectors +78 points [+50, +143]. Seed 1 is +65 points from the average, so the gap does not depend on the seed. The two joint models still differ in other ways: 4.0 points of multilingual accuracy and 3.6 points of ECE. The strongest merge that is not tuned on the test items and keeps general accuracy is TIES at its usual setting (keep 20%, lambda 1): it keeps 64%, +56 points [+40, +97] from joint training with seed 0. Ensembles of the specialists' predictions do not reach joint training either. Averaging probabilities keeps 26% of the gains with v1 as a member, geometric averaging 36% at best, and choosing the most confident member 54%, which is about what the weight average keeps (52%) at four or five forward passes per call. A skill that only one member has is easily outvoted when the members are averaged; picking the most confident member lets that member speak when it is sure (our reading, not tested separately).

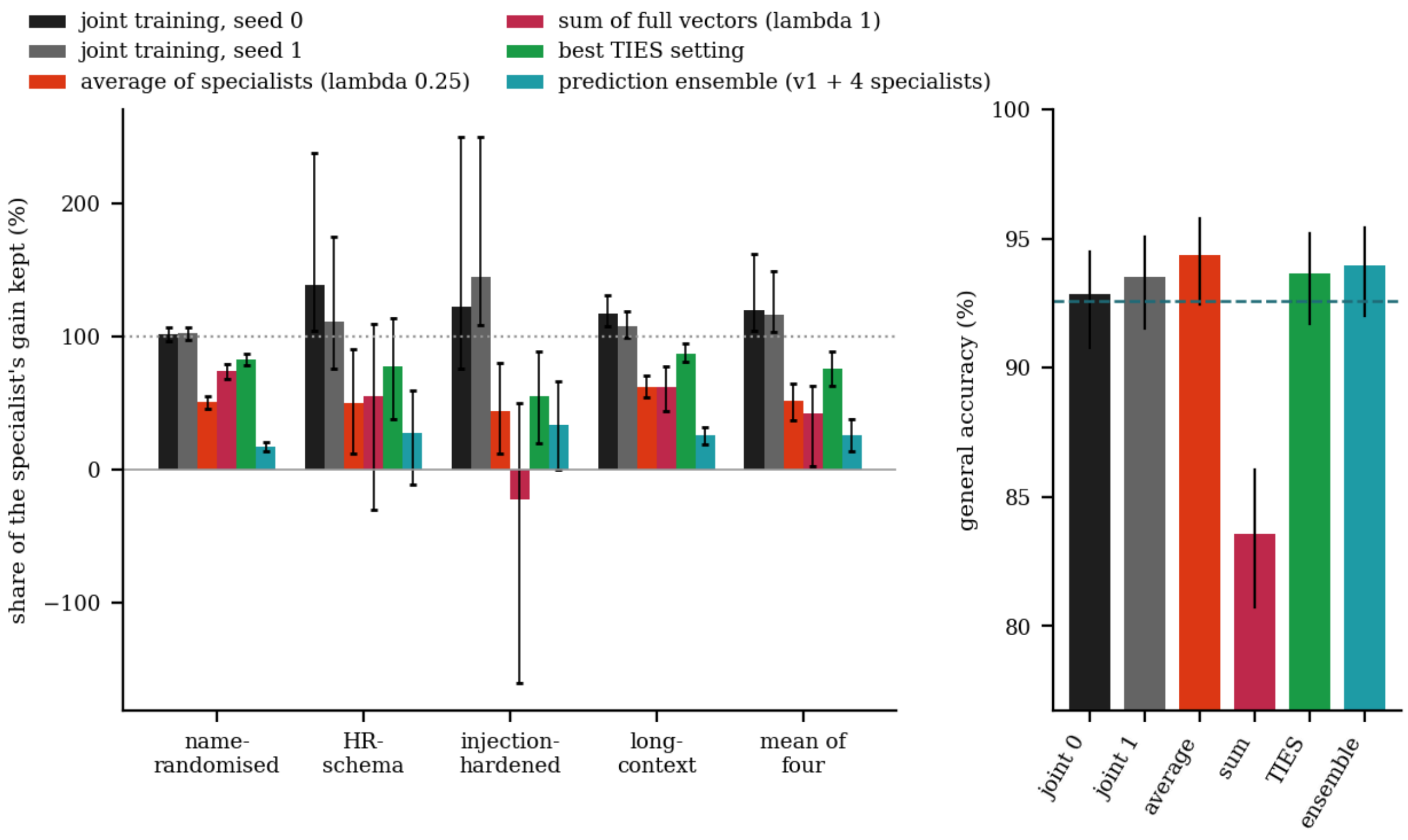

**Figure 18.** Joint training against the merges. Left: share of each specialist's gain kept, with bootstrap intervals; 100% is the specialist itself. Right: general accuracy (Wilson interval; dashed: v1). The joint model is one fine-tune on the union of the v1 rows and the extra rows of the four specialists; the ensemble averages the output probabilities of five models and is not one set of weights.

**Table 16.** Joint training against merging and ensembling. Mean gain kept over the four specialist skills [95% CI], general accuracy and its change against v1 (points), multilingual accuracy (the training rows include some non-English text) and ECE (%). Intervals are bootstrap over test items; they do not include training-seed variation, which the two joint rows show directly. Every row, the joint models included, uses v1's stored temperature, so ECE compares weights and not calibration fits.

| Model | General (%) | General vs v1 (points) | Mean gain kept (%) | Multilingual (%) | ECE (%) | Cost |
|---|---|---|---|---|---|---|
| v1 (plain fine-tune) | 92.6 [90.5-94.3] | - | 0 by definition | 73.6 [68.5-78.1] | 4.0 | |
| mean of the four specialists | 93.0 | +0.4 | 100 by definition | 77.9 | 3.8 | four models |
| joint fine-tune, seed 0 | 92.9 [90.8-94.5] | +0.3 [-1.3, +2.2] | 120 [104, 162] | 77.3 [72.4-81.6] | 5.0 | one set of weights, trained on all rows |
| joint fine-tune, seed 1 | 93.5 [91.5-95.1] | +0.9 [-0.7, +2.7] | 116 [104, 149] | 73.3 [68.2-77.8] | 1.4 | one set of weights, trained on all rows |
| weight average of the four specialists, lambda 0.25 | 94.3 [92.4-95.8] | +1.7 [+0.3, +3.4] | 52 [38, 64] | 78.0 [73.1-82.1] | 2.6 | free |
| weight sum of the four vectors, lambda 1 | 83.6 [80.7-86.1] | -9.0 [-11.7, -6.2] | 42 [3, 63] | 67.4 [62.1-72.3] | 10.3 | free |
| TIES from v1, keep 20%, lambda 1 (the method's usual setting) | 94.1 [92.1-95.6] | +1.5 [-0.3, +3.2] | 64 [50, 76] | 77.3 [72.4-81.6] | 3.2 | free |
| best tIES from v1, keep 20%, lambda 1.5 (picked on the test items) | 93.7 [91.7-95.2] | +1.1 [-0.8, +2.8] | 76 [63, 89] | 76.7 [71.8-81.0] | 4.0 | optimistic |
| DARE p 0.5, lambda 1 (mean of 3 masks) | 83.0 | -9.6 (range -10.1 to -9.3) | 44 (range 40 to 53) | 67.3 | 10.5 | free |
| DARE p 0.9, lambda 1 (mean of 3 masks) | 83.1 | -9.5 (range -9.8 to -9.3) | 38 (range 33 to 40) | 66.4 | 9.9 | free |
| prediction ensemble, mean of probabilities: v1 + 4 specialists | 93.9 [92.0-95.4] | +1.3 [+0.0, +2.8] | 26 [14, 38] | 81.1 [76.4-85.0] | 1.5 | 5 forward passes |
| prediction ensemble, mean of probabilities: 4 specialists | 94.6 [92.8-96.0] | +2.0 [+0.5, +3.8] | 28 [10, 40] | 80.1 [75.4-84.1] | 1.5 | 4 forward passes |
| prediction ensemble, geometric mean: v1 + 4 specialists | 94.1 [92.1-95.6] | +1.5 [+0.1, +3.1] | 26 [15, 38] | 80.1 [75.4-84.1] | 2.3 | 5 forward passes |
| prediction ensemble, geometric mean: 4 specialists | 94.9 [93.1-96.3] | +2.3 [+0.7, +4.0] | 36 [21, 49] | 80.1 [75.4-84.1] | 2.2 | 4 forward passes |
| prediction ensemble, most confident member: v1 + 4 specialists | 94.5 [92.6-95.9] | +1.9 [+0.4, +3.5] | 54 [42, 73] | 78.0 [73.1-82.1] | 4.4 | 5 forward passes |
| prediction ensemble, most confident member: 4 specialists | 95.0 [93.2-96.4] | +2.4 [+0.8, +4.0] | 54 [43, 74] | 78.6 [73.8-82.7] | 4.0 | 4 forward passes |

Figure 19 puts the same models on the plane of Figure 17.

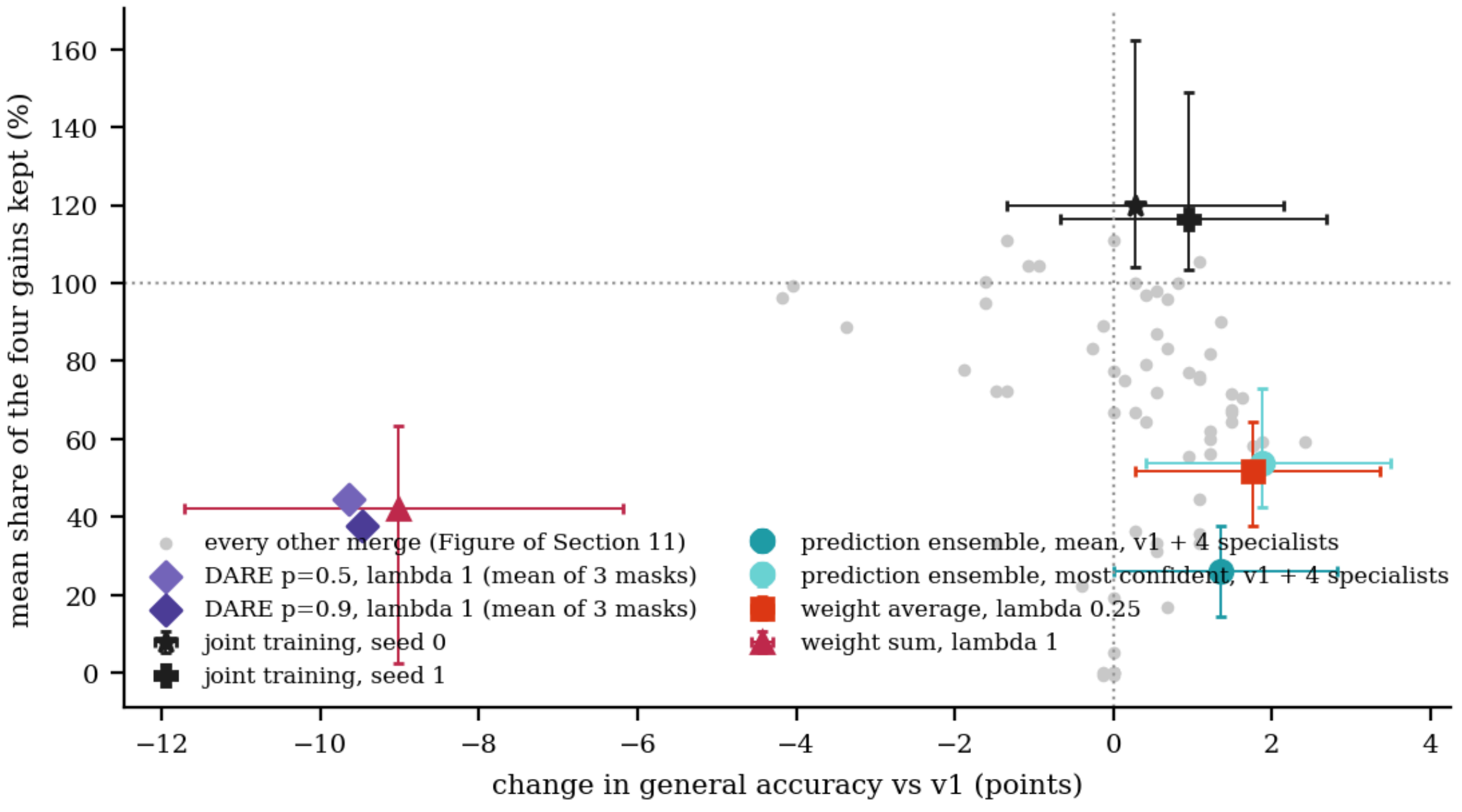

**Figure 19.** Where joint training sits among the merges: change in general accuracy against v1 (x) and mean share of the four gains kept (y), with bootstrap intervals for the headline models. Grey dots are the other merged models of this study.

The weight-space picture does not make the joint model a point on the merge lines. It differs from v1 by 9.18 in Frobenius norm, against 6.71 between two plain seeds, and it differs from the weight average of the four specialists by 8.30, which is 1.24 times a seed difference. Measured from v1, the cosine between the joint model's change and the mean of the four specialist changes is 0.45; for a plain seed it is 0.60, which is higher only because every vector measured from v1 carries the same minus-v1 term. Measured from the released weights, where that term is absent, the joint model's change has cosine 0.47 with the mean specialist change, plain seeds have 0.54 on average and the specialists themselves 0.75. So the joint model sits no nearer to the specialist average than an independent plain run does. That fits the accuracy gap above but does not explain it (Figure 20).

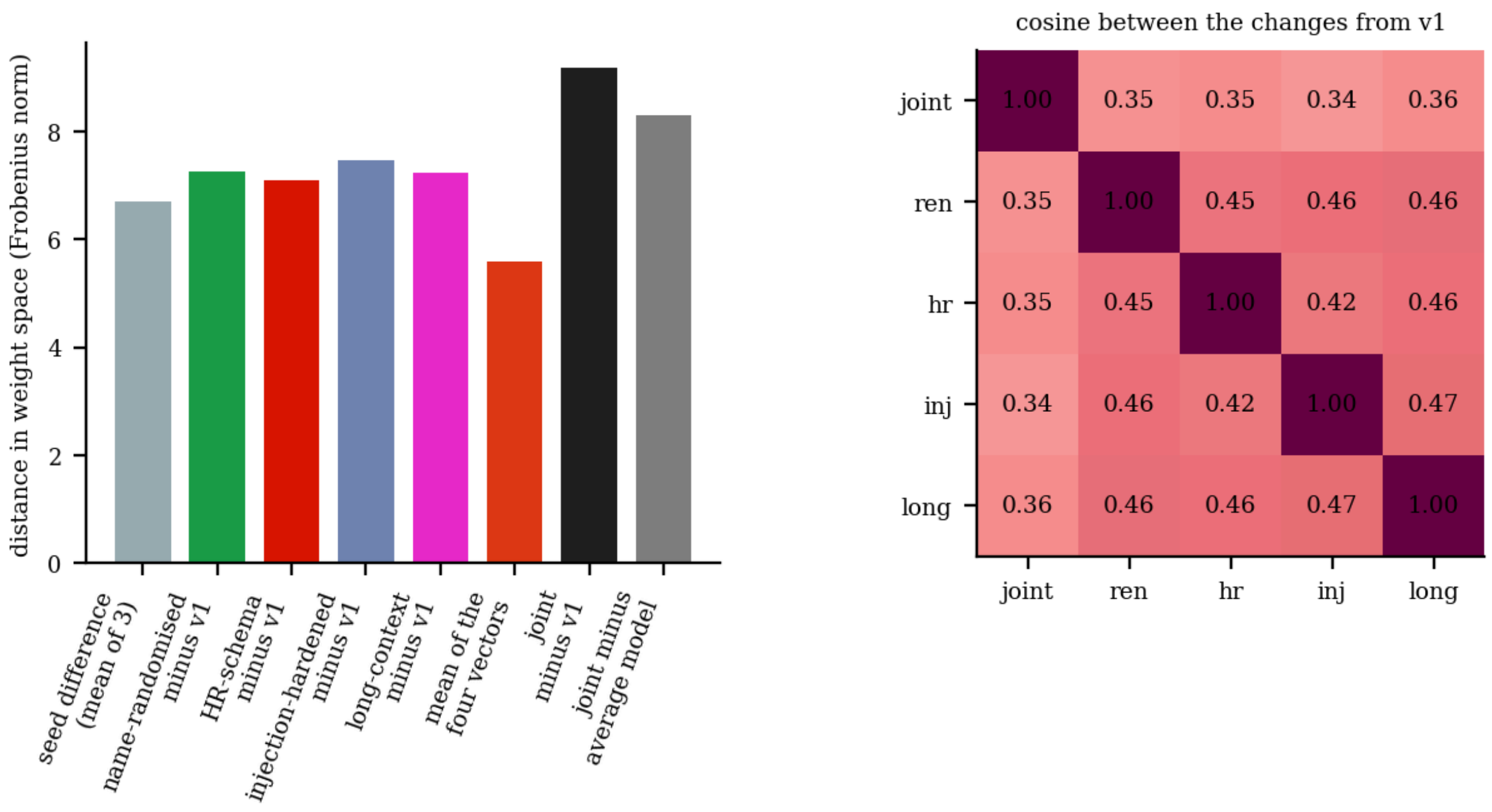

**Figure 20.** Where the joint model sits in weight space. Left: distances between checkpoints; the joint model is about as far from the average of the specialists as from v1. Right: cosine between the change from v1 to the joint model and the changes from v1 to each specialist.

## 13 Random drop-and-rescale (DARE)

DARE (Yu et al., arXiv:2311.03099) sets a random share $p$ of each task vector to zero and rescales the rest by $1/(1-p)$, on the argument that fine-tuning changes are highly redundant, and then merges the thinned vectors. We apply it to the four specialist vectors (each minus v1) with three random masks per setting. It tests the claim that interference between vectors can be removed by thinning them; if it can, the sum at $\lambda = 1$, which loses general accuracy, should recover. At $\lambda = 1$ and $p = 0.5$, general accuracy changes by -9.6 points against v1 (plain sum: -9.0) and the mean gain kept is 44% (plain sum: 42%). At $\lambda = 1$ and $p = 0.9$, general accuracy changes by -9.5 points against v1 (plain sum: -9.0) and the mean gain kept is 38% (plain sum: 42%). Without the rescale at $p = 0.9$ the mean gain kept is 22%.

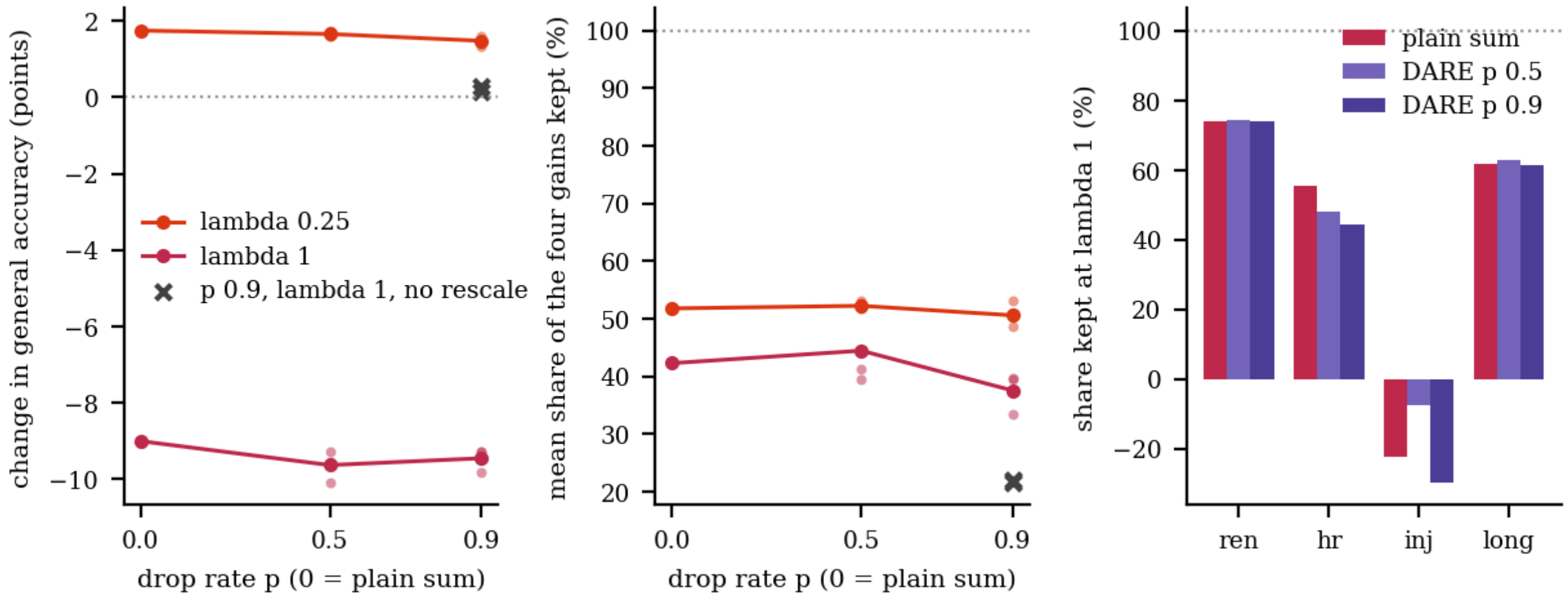

**Figure 21.** Random drop and rescale (DARE) applied to the four specialist vectors before summing. Left and middle: change in general accuracy and mean gain kept against the drop rate p; dots are mask seeds, lines their mean; p = 0 is the plain sum. Right: gain kept by skill at lambda 1. The crosses are the same p = 0.9 without the 1/(1-p) rescale.

**Table 17.** DARE on the four specialist vectors (rooted at v1). Weight per vector lambda, drop rate p, number of mask seeds, change in general accuracy against v1 in points (mean, range over mask seeds), mean gain kept over the four skills in % (mean, range) and decisions flipped against v1.

| lambda | Drop rate p | Masks | General vs v1 (points) | Mean gain kept (%) | Flipped (%) |
|---|---|---|---|---|---|
| 0.25 | 0 (plain sum) | - | +1.7 | 52 | |
| 0.25 | 0.5 (rescaled) | 3 | +1.7 (+1.6 to +1.7) | 52 (52 to 53) | 9.0 |
| 0.25 | 0.9 (rescaled) | 3 | +1.5 (+1.3 to +1.6) | 51 (49 to 53) | 9.0 |
| 1 | 0 (plain sum) | - | -9.0 | 42 | |
| 1 | 0.5 (rescaled) | 3 | -9.6 (-10.1 to -9.3) | 44 (40 to 53) | 24.7 |
| 1 | 0.9 (rescaled) | 3 | -9.5 (-9.8 to -9.3) | 38 (33 to 40) | 25.0 |
| 1 | 0.9 (no rescale) | 2 | +0.2 (+0.1 to +0.3) | 22 (22 to 22) | 3.6 |

## 14 Does the headline depend on one seed?

Three of the five recipes have a single seed, so a headline merge could be a feature of one lucky or unlucky run. The rename recipe has three seeds, and we rebuilt the four-way merge with each of them. Across seeds the mean gain kept at $\lambda = 0.25$ spans 8 points and at $\lambda = 1$ it spans 23 points (Table 18). The intervals overlap, so one seed does not prove the others wrong, but a single merged model, and the full sum above all, is a noisy number: the general accuracy change at $\lambda = 1$ runs from -10.1 to -6.3 points.

**Table 18.** Does the four-way merge depend on which seed of the rename specialist is used? Mean gain kept over the four skills (the rename skill is scored against the rename specialist of seed 0 in every row) and the change in general accuracy against v1.

| lambda | Rename vector from | Mean gain kept (%) [95% CI] | Rename gain kept (%) | General vs v1 (points) |
|---|---|---|---|---|

| lambda | Rename vector from | Mean gain kept (%) [95% CI] | Rename gain kept (%) | General vs v1 (points) |
|---|---|---|---|---|
| 0.25 | rename seed 0 (the one used above) | 52 [38, 64] | 51 | +1.7 [+0.3, +3.4] |
| 0.25 | rename seed 1 | 55 [39, 69] | 52 | +1.7 [+0.3, +3.4] |
| 0.25 | rename seed 2 | 46 [32, 58] | 51 | +1.2 [-0.3, +2.8] |
| 1 | rename seed 0 (the one used above) | 42 [3, 63] | 74 | -9.0 [-11.7, -6.2] |
| 1 | rename seed 1 | 57 [25, 83] | 79 | -10.1 [-13.1, -7.1] |
| 1 | rename seed 2 | 66 [33, 92] | 76 | -6.3 [-9.2, -3.8] |

## 15 Held-out shifts

The specialist skills of Section 3 are in-family: each is measured on held-out items of the kind the specialist was trained on. A merge that keeps those skills could still have lost something elsewhere, so we also scored the headline models on two shifts that differ from the training files. Public benchmark questions (ARC-Easy, ARC-Challenge, MMLU and HellaSwag, 200 items each) are a different task and answer format from the ticket decisions in training, and the released base is near chance on them; they therefore say little about merging unless a merge falls below the base. The code-mixed tickets (Hinglish, Arabizi and Spanglish; 156 items) keep the department-routing schema of the training rows but write the text in a mix of languages and scripts that none of the training categories covers; the training rows do contain monolingual non-English text, so this is a shift in text and not in task. We had also scored a set of tickets from a2-style held-out schemas, but both of its label sets occur in the joint training file and its categories are those of the training rows, so it is not a shift and we drop it.

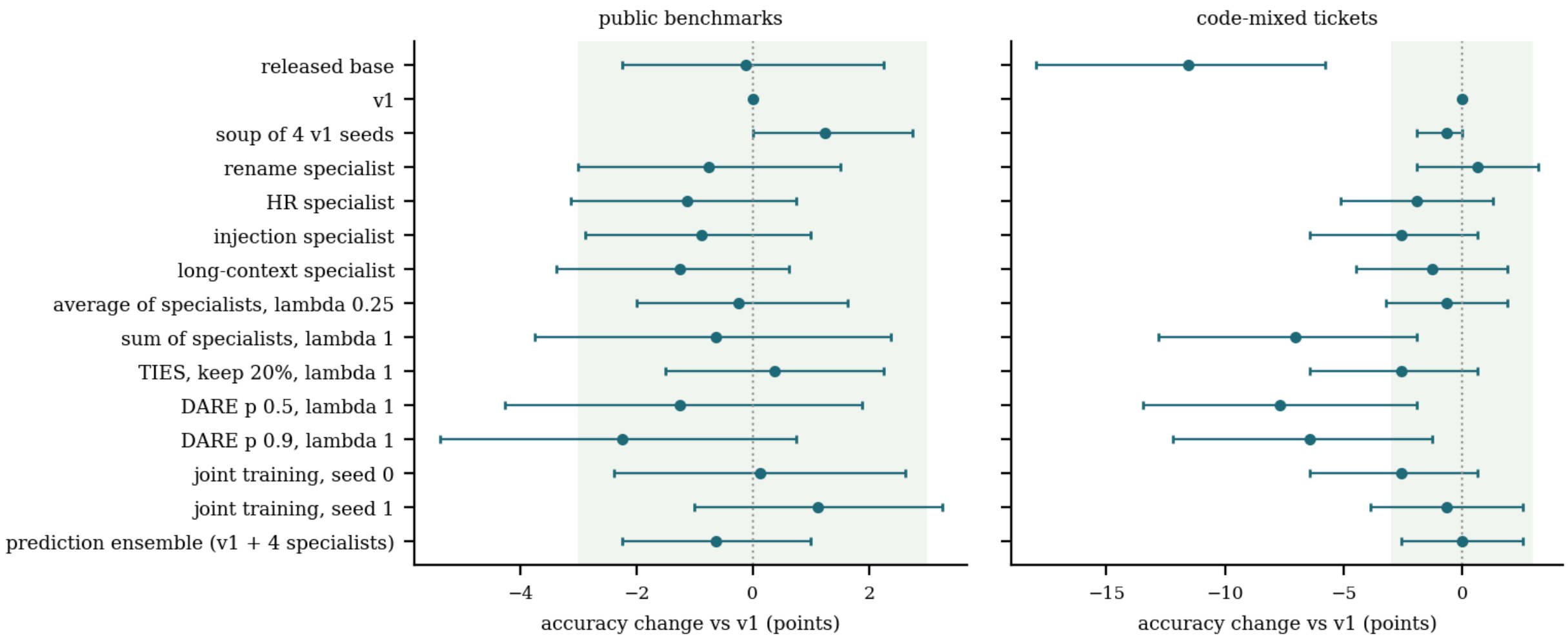

**Figure 22.** Accuracy change against v1 on two shifts, with paired bootstrap intervals. The green band is plus or minus 3 points. Public benchmark questions: 800 items; code-mixed tickets: 156 items.

**Table 19.** Accuracy (%) on two held-out shifts and, in brackets, the change against v1 in points. No item of these sets appears in any training file (exact text match).

| Model | public benchmarks | code-mixed tickets |
|---|---|---|
| released base | 36.4 (-0.1) | 84.6 (-11.5) |
| v1 | 36.5 (+0.0) | 96.2 (+0.0) |
| soup of 4 v1 seeds | 37.8 (+1.3) | 95.5 (-0.6) |
| rename specialist | 35.8 (-0.8) | 96.8 (+0.6) |
| HR specialist | 35.4 (-1.1) | 94.2 (-1.9) |
| injection specialist | 35.6 (-0.9) | 93.6 (-2.6) |
| long-context specialist | 35.2 (-1.3) | 94.9 (-1.3) |

| Model | public benchmarks | code-mixed tickets |
|---|---|---|
| average of specialists, lambda 0.25 | 36.2 (-0.3) | 95.5 (-0.6) |
| sum of specialists, lambda 1 | 35.9 (-0.6) | 89.1 (-7.1) |
| TIES, keep 20%, lambda 1 | 36.9 (+0.4) | 93.6 (-2.6) |
| DARE p 0.5, lambda 1 | 35.2 (-1.3) | 88.5 (-7.7) |
| DARE p 0.9, lambda 1 | 34.2 (-2.2) | 89.7 (-6.4) |
| joint training, seed 0 | 36.6 (+0.1) | 93.6 (-2.6) |
| joint training, seed 1 | 37.6 (+1.1) | 95.5 (-0.6) |
| prediction ensemble (v1 + 4 specialists) | 35.9 (-0.6) | 96.2 (+0.0) |

The four-way weight average, the soup, the joint models and the ensemble stay within 2.6 points of v1 on both shifts, which is inside the sampling noise of a few hundred items (Figure 22); we see no out-of-family cost of merging for them. The full-vector merges are different: the worst of them (DARE p 0.5, lambda 1) loses 7.7 points on code-mixed tickets, which matches their loss of general accuracy in Section 7. The benchmark questions separate nothing: the released base scores 36.4% on them and v1 36.5%. Joint training (seed 0) and the four-way weight average differ by +0.4 points on public benchmarks, -1.9 points on code-mixed tickets.

## 16 Scorecard of the merging theories

The methods tested here are published; what we can add is evidence about which of the explanations given for them hold in this setting. We turned each explanation into a prediction with a numerical threshold before running the new experiments, and Table 20 applies the thresholds to what we measured. Of the 10 predictions, 2 were confirmed (or partly), 7 were refuted and 1 cannot be tested with this collection. Several of the tests are weak: a single base, two joint seeds, six specialist pairs, four skills. A refuted prediction means that it did not appear in these data, not that the theory is wrong elsewhere. One prediction was deliberately the opposite of the naive reading of its theory: the equivalence of a task-vector sum and a multitask gradient step is exact only for one epoch of full-batch gradient descent, so we predicted that it would break after four epochs of mini-batches.

**Table 20.** Scorecard of the merging theories we tested. The prediction and its threshold were written down before the new runs; the last column applies the threshold to the measurement.

| Theory | Source | Prediction fixed in advance | Measured | Outcome |
|---|---|---|---|---|
| Task vectors are multitask gradients, so the specialist average should approximate joint training (ATM; exact only for one epoch of full-batch gradient descent, whereas we train four epochs of mini-batches) | arXiv:2411.03055 | the approximation breaks: joint training keeps at least 10 points more than the four-way average | joint minus average: seed 0 +68 points [+51, +108]; seed 1 +65 points [+47, +98] | confirmed |
| A specialist vector is portable to another seed of the same base (task arithmetic) | arXiv:2212.04089 | each of the four vectors installs at least 70% of its gain in seeds 1 and 2 | ren 79%, hr 64%, inj 61%, long 103% | partly confirmed |
| Delta parameters are redundant, so random drop-and-rescale removes interference (DARE) | arXiv:2311.03099 | at lambda 1 DARE recovers at least 3 points of general accuracy against the plain sum | best recovery -0.4 points; mean gain kept 38% at p 0.9 against 42% for the plain sum | refuted |

| Theory | Source | Prediction fixed in advance | Measured | Outcome |
|---|---|---|---|---|
| Merging loses to prediction averaging because weight space is the bottleneck (soups, RegMean comparisons) | arXiv:2203.05482; arXiv:2212.09849 | the best of six prediction ensembles keeps at least 10 points more than the best weight merge | best ensemble (most confident member: 4 specialists) 54% against best merge 52% (ensemble minus average: +2 points [-10, +19]); the geometric-mean and most-confident variants were added after the plain mean missed the threshold | refuted |
| Orthogonal task vectors merge without interference (task-feature specialization) | arXiv:2604.17078 | across the six specialist pairs, \|cosine\| and 50/50-average retention have a rank correlation of at most -0.5 | rank correlation -0.14 [-1.00, +0.80], n = 6; cosines span only 0.42 to 0.47 | not testable here: the vectors are all about equally non-orthogonal |
| A skill the base model already has merges better (mergeability follows the base model's prior knowledge) | arXiv:2601.06672 | rank correlation between the base's accuracy on a skill and the gain kept by the four-way average is positive and at least 0.5 | rank correlation +0.00, n = 4 skills | refuted |
| The barrier along a line grows with the squared distance between its ends (quadratic-bowl picture of linear mode connectivity) | arXiv:1912.05671 | rank correlation between squared length and midpoint barrier of at least 0.6 over all lines with an evaluated midpoint | rank correlation -0.34 [-0.69, +0.12], 22 lines | refuted |
| The headline merge is not a feature of one seed (soups) | arXiv:2203.05482 | swapping the rename vector for its seed 1 or seed 2 changes mean gain kept by less than 5 points | spread over the three seeds: 8 points at lambda 0.25, 23 points at lambda 1 | refuted |
| The joint model is closer to the specialist average than a seed difference is (geometry of task vectors as gradients) | arXiv:2411.03055 | distance(joint, average) / seed difference below 1.5, and cosine(joint change, mean specialist change) above 0.5 | distance ratio 1.24; cosine 0.45 (a plain seed's change gives 0.60) | refuted |
| Merging neither adds nor removes ability on distributions nobody trained on (held-out shifts) | this study | every merge, ensemble and the joint model is within 3 points of v1 on each of the two shifts | largest change 7.7 points (DARE p 0.5, lambda 1, code-mixed tickets); gentle merges, joint models and ensemble stay within 2.6 | refuted |

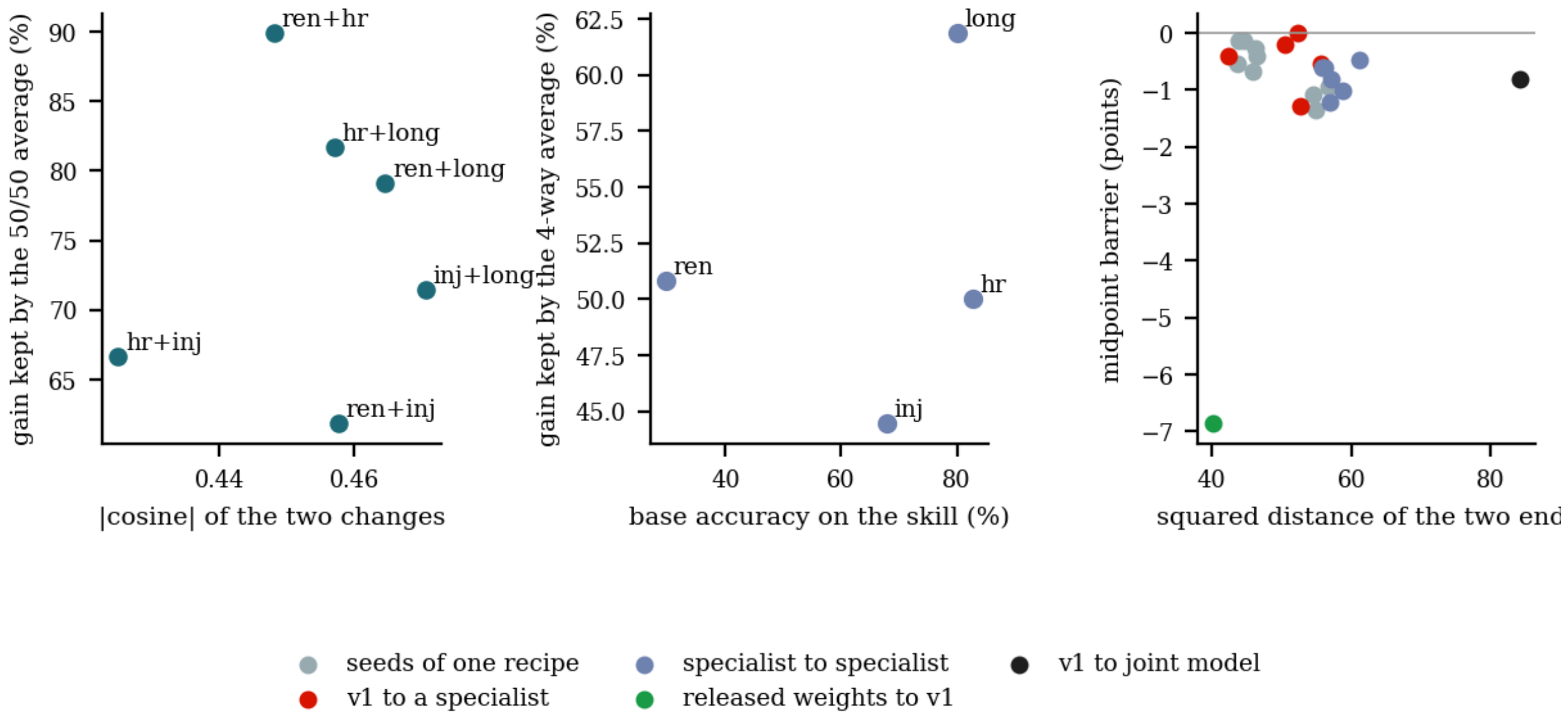

**Figure 23.** Three tests from stored results. Left: if orthogonal vectors merged cleanly, points would fall on a falling line, but all cosines are about 0.45. Middle: gain kept by the four-way average against the released base's own accuracy on each skill. Right: midpoint barrier in general accuracy against the squared distance between the two ends, for every pair of models with an evaluated 50/50 point.

## 17 Cost, and what was not run

### 17.1 What a merge costs compared with retraining

A four-epoch fine-tune of one specialist takes about five minutes on one RTX 4090 when the card is free. Building a merged state dict takes seconds on the CPU, and evaluating one merged model on the whole suite took about a minute on a GPU that was shared with other jobs; 156 models were evaluated this way. The joint fine-tune of Section 12 trains on 5,544 rows instead of 2,818; on our shared card, with the optimizer state kept on the CPU to stay under 6 GB of GPU memory, it took under 30 minutes of wall-clock time per seed (one run included a crash of the shared GPU and a resume from the state saved after epoch 2). In rows read per epoch, the four specialists together use 13,998 and one joint model 5,544, so a joint run costs about 2.0 plain runs; its price is that it must be redone whenever a specialist is added, which a merge avoids. A merge needs neither the data nor the trainer, only the weights, but it needs an evaluation suite to find out which merge is good, and in this study that suite is what takes the time. If a team already has the suite, merging many candidates is cheaper than retraining; if it has not, a merge is a blind guess. A prediction ensemble costs one forward pass per member at every call, which is what a merge is meant to avoid.

### 17.2 What was not run

The joint fine-tune was trained with two seeds, which is too few to put a tight interval on seed-to-seed variation of joint training; the plain-seed spread of Section 4 is the other guide. We did not run a sequential fine-tune (train on one specialist after another), which is the other alternative to merging and is known to forget; the TIES grid was cut to five settings and the layer masks to four of seven per specialist (Sections 9 and 10); we did not run Fisher-weighted averaging, RegMean, learned merge weights or permutation alignment; we did not test whether rounding merged weights to half precision matters (16% of coordinates of a fine-tune are unchanged because of half-precision storage, which the merge arithmetic in 32 bits does not see). The same holds for merging models with a different layout of the input sequence, which we did not attempt.

## 18 Limitations and threats to validity

- **LLM-made labels.** All training and test labels were produced by one language model and filtered by its own blind second pass. A second model agreed with the stored label on 94-99% of held-out items (Section 2.2), which bounds how much label noise can affect the comparisons, but it is not a human gold standard. A merge that exactly reproduces the labeler's quirks scores well.

- **Synthetic data.** Cases are generated, not collected. Specialist skills are therefore skills on generated variants of the problem; none of the results here says anything about real tickets or real prompt injections.
- **One shared base.** Every specialist starts from the same released weights and shares most of its rows with v1. This is the best case for merging (all models in one basin), and it also means the specialists differ from v1 only a little. Fine-tunes started from different bases, trained for longer, or trained with adapters could behave differently.
- **Few seeds.** There are four plain seeds, three seeds for two specialists and one seed for the other three. A specialist seen once cannot separate a recipe effect from seed noise; the seed analysis in Section 8, the seed-swap test in Section 14 and the transfer test in Section 7.1 are the controls we have.
- **Selection on test items.** The sweep over $\lambda$, trim density and mask was evaluated on the same held-out sets that produce the numbers. The rows of the recipe table that pick the best setting are upper bounds; a deployment would need a separate selection set.
- **Item overlap and multiple comparisons.** The evaluation items overlap across models, so all comparisons are paired, but we report about 156 models with intervals at the 95% level and made no multiplicity correction. A single interval that excludes zero among many should not be read as a finding.
- **Half-precision storage and fixed temperature.** The released weights and every fine-tune are stored in half precision; merges are computed in 32-bit and loaded back, and every merged model uses v1's stored temperature. ECE values therefore describe an uncalibrated merge, not the best one achievable.
- **Low-gain specialists.** The HR specialist improves on v1 by only a few points, so its gain-kept share has a wide interval; the no-shuffle model is worse than v1 on its own skill, so no share is reported for it.
- **Simple baselines and variants.** The prediction ensemble is the plain mean of probabilities from five models; a weighted or routed ensemble (an oracle that knew which specialist to ask) would keep more. DARE is applied to specialist-minus-v1 vectors with two drop rates and three masks, not to specialist-minus-base vectors as in its original merges. The joint model sees the union of the data, so it is the right reference for what the data allow, not a compute-matched rival of one specialist.
- **Two joint runs.** The joint fine-tune and its comparison with the merges rest on two training seeds. Two seeds say whether joint training is stable but cannot bound its variance well, and the intervals in Section 12 do not include seed variation.
- **Thresholds of the scorecard.** The numerical thresholds of Section 16 were chosen by us before the runs and are not derived from the theories; a different threshold could flip a verdict near the boundary, and several tests rest on four skills or six pairs.
- **Held-out shifts.** The three shifts are only some of the shifts one could ask about; public benchmark questions are far from the training task, so a flat result there may say the merged encoders are simply near chance on them.
- **Unexamined methods.** We did not test Fisher-weighted averaging, learned merge weights, permutation alignment, adapter merging or any method that needs training data; a result here is a statement about the methods tried (soups, task arithmetic, TIES, DARE, prediction ensembling).

## 19 References and related work

**Weight averaging and linear connectivity.** Frankle et al. (arXiv:1912.05671) showed that networks fine-tuned from the same early checkpoint can be joined by a straight line without an accuracy barrier, which is the property behind our seed-line test. Wortsman et al. (arXiv:2203.05482) averaged the weights of fine-tunes of one pre-trained model and found the average at least as good as the members; our soups repeat that design on a small non-autoregressive encoder. Matena and Raffel (arXiv:2111.09832) weight each parameter by its Fisher information instead of using a uniform mean; we did not test that. Jin et al. (arXiv:2212.09849) merge fine-tuned language-model classifiers with weights chosen to keep each model's predictions on its own data, a method that needs activation statistics from every task; Choshen et al. (arXiv:2204.03044) average fine-tunes of one base to build a better starting point, a use we did not test.

**Task arithmetic and TIES.** Ilharco et al. (arXiv:2212.04089) define task vectors as fine-tune minus pre-trained weights and show that adding, scaling and negating them edits behaviour; Yadav et al. (arXiv:2306.01708) trim small entries and resolve sign conflicts before merging. We implemented both on the same checkpoints and report their outcome as a recipe table. Rahamim et al. (arXiv:2601.06672) argue that merging succeeds more often when the base model already knows the fine-tuning data, and Hitit et al. (arXiv:2511.21437) study merging experts with overlapping or conflicting objectives. Our setting is a case of the second kind in a small way: the specialists overlap in training rows. Layer-aware task arithmetic (Chen et al., arXiv:2502.20186) gives each layer its own task-vector weight; our layer masks are the crudest version of that idea, a 0 or 1 choice per block of layers.

**Surveys and recent merging studies.** Song and Zheng (arXiv:2603.09938) survey merging for language models, from weight averaging and task vectors to sparsification and mixtures of experts, and give the taxonomy we use informally. Choudhary et al. (arXiv:2607.27240) merge two Gemma-3-1B-IT fine-tunes, one for harm classification and one for refusal, with linear, SLERP, TIES and DARE-TIES; they find that the task vectors are nearly orthogonal and that the fine-tune with the larger per-layer magnitudes wins, which is the kind of scale effect we measure for our specialists. Tang et al. (arXiv:2606.19549) predict whether LoRA updates will merge from early training signals; we did not train adapters or predict mergeability, and our norm and cosine tables are a post-hoc version of the same idea for full fine-tunes.

**Explanations of why merging works, and where they were tested.** Zhou et al. (arXiv:2411.03055) show that under single-epoch full-batch gradient descent a task vector is the scaled negative gradient of its task loss, so the sum of task vectors is one step on the sum of the task losses; their experiments are on vision tasks. We test the consequence for our encoder by comparing the merge with a joint fine-tune (Section 12). Yu et al. (arXiv:2311.03099) argue that fine-tuning changes are highly redundant, drop a random share of each vector and rescale the rest, and report that this lets homologous models merge with less interference, including for encoder models on GLUE; Section 13 applies it to our specialists. Ortiz-Jimenez et al. (arXiv:2305.12827) name weight disentanglement as the property that makes task arithmetic work, and Liu et al. (arXiv:2604.17078) tie it to task-feature specialization and orthogonal weight updates; we test the orthogonality prediction on six specialist pairs in Section 16. Yadav et al. (arXiv:2410.03617) study averaging, task arithmetic, DARE and TIES for models of 1B to 64B parameters on held-in and held-out tasks, and He et al. (arXiv:2505.10833) benchmark merging of domain-specialized language models of 2B to 9B parameters; both are at a scale far from ours, and neither has a typed-decision encoder. Ramé et al. (arXiv:2205.09739) connect weight averaging to functional ensembling through a bias-variance-covariance-locality decomposition, which is the reason we include a prediction ensemble as a baseline.

**Typed-decision models.** Sharma et al. (arXiv:2610.11392) introduce a benchmark for calibration, framing sensitivity and cost in models that return probabilities over typed answers; Zhang et al. (arXiv:2610.00831) read typed decisions out of an instruction-tuned language model and correct label and position priors without any parameter change; Geuter et al. (arXiv:2610.17205) fine-tune open language models with LoRA and a decision head. None of these works merges or averages fine-tunes of such a model.

**What is and is not new here.** Soups, task arithmetic, TIES and linear connectivity are published methods and findings, and we claim none of them. We did a title-level search of the Hugging Face paper index for merging, soups and weight averaging with decision-model terms and did not find a study that applies them to a typed-decision model; the search was not exhaustive. What we add is a test of the explanations given for those methods on one model: a family of seeds and five specialists on one base, the geometry of their changes computed exactly from the stored weights, a joint fine-tune and a prediction ensemble as the baselines a merge should be judged against, DARE, and a scorecard of predictions written down before the runs with the ones that failed reported. We make no claim about larger models, other bases, adapters or about other kinds of skill.

## 20 Reproducibility

**Code and data.** Everything is in the project directory. `a11_run.py weights` computes the per-layer inner-product matrices and sign statistics of the task vectors (output: `results/a11_weights.json`, SHA-256 starts `529f52ffe5da5e78`); `a11_run.py eval` builds each merged model in memory, evaluates it on the suite of Section 3 and writes one file `results/a11_ev__<merge>.json` (currently 157); the evaluation resumes where it stopped. The builder `lp_u11.py` reads those files, the training row files and the stored model weights' statistics, computes every number and figure in this document, and prints it. The software versions were laya 0.4.26, torch 2.11.0+cu128, numpy 2.2.6; evaluation used one NVIDIA GPU shared with other jobs.

**Determinism.** Merges are exactly reproducible (the arithmetic is deterministic on the CPU); GPU evaluation is not bit-reproducible, and two evaluations of identical weights differ on 0.00% of decisions (Section 3.2). Bootstrap resamples use fixed seeds.

**Scope of what is stored.** The merged models are not saved, only their per-item predictions; that is enough to recompute every statistic here. The specialist checkpoints are the ones listed in Table 1 and were trained with Laya's own trainer from the released weights. The two joint models are stored (`models/a11_joint_s0`, `models/a11_joint_s1`).

**Second-round scripts.** `a11_train.py train <seed>` trains the joint model with Laya's trainer on `data/a11_joint.jsonl` (the v1 rows plus the extra rows of the four specialists); the AdamW state is kept on the CPU so the run stays under 6 GB of GPU memory, and the loop is the library's own with one addition, a full training state saved after every epoch so that a crash of the shared GPU does not lose the run. Seed 0 crashed once at the start of epoch 3 and resumed from that state; seed 1 crashed once in epoch 1, before any state existed, and restarted from the beginning. `a11_run2.py eval` evaluates the joint models, cross-seed transfers, rename-seed swaps and DARE merges; `a11_ens.py` records every model's probabilities once and builds the ensembles by replaying them (replaying one member alone reproduces it exactly: general accuracy 93.5% against 93.5%, 0 items differ); `a11_shift.py` scores the held-out shifts; `a11_run2.py geom` writes the weight geometry of the joint models. The predictions of the second round are fixed in `reviews/a11_ideas.md`, with its time stamp.

The text was drafted and edited with LLM assistance; all numbers are recomputed from the stored results by the build scripts and were checked by the authors.

## Appendix A  Every evaluated model

**Table 21.** Every evaluated model: general accuracy (six English sets), multilingual, swap-name, HR, injection, long-filler accuracy, order stability (where computed), ECE and decisions flipped against v1. All values in %. Tags are the merge descriptions used by the driver (lerp a b alpha; soup of members; sum = v1 + lambda times the listed specialist vectors; ties = root, density, lambda; mask = specialist, part moved).

| Model | Gen. | ML | Swap | HR | Inj. | Long | Order | ECE | Flip |
|---|---|---|---|---|---|---|---|---|---|
| m · a11 joint s0 | 92.9 | 77.3 | 83.1 | 93.4 | 97.2 | 95.9 | 91.3 | 5.0 | 10.7 |
| m · a11 joint s1 | 93.5 | 73.3 | 83.5 | 91.6 | 99.1 | 93.4 | 93.6 | 1.4 | 10.0 |
| m · base | 72.9 | 61.2 | 29.7 | 82.8 | 67.9 | 80.0 | 74.2 | 29.6 | 26.7 |
| m · v1 | 92.6 | 73.6 | 32.7 | 84.2 | 86.8 | 66.4 | 92.8 | 4.0 | 0.0 |
| m · v1 hr | 93.5 | 78.0 | 44.0 | 90.8 | 89.6 | 77.0 | 93.0 | 3.4 | 11.5 |
| m · v1 inject | 93.7 | 76.7 | 26.2 | 83.9 | 95.3 | 66.2 | 93.2 | 4.0 | 10.1 |
| m · v1 long | 92.9 | 78.3 | 35.8 | 84.2 | 87.7 | 91.5 | 92.1 | 3.8 | 11.8 |
| m · v1 noshuf | 92.1 | 74.2 | 49.0 | 82.4 | 84.9 | 63.7 | 88.6 | 4.0 | 9.0 |
| m · v1 noshuf s1 | 92.1 | 75.5 | 47.4 | 84.6 | 85.8 | 67.3 | 86.2 | 4.1 | 9.1 |
| m · v1 noshuf s2 | 91.8 | 74.2 | 54.0 | 84.6 | 80.2 | 57.2 | 91.4 | 2.8 | 8.6 |
| m · v1 rename | 91.9 | 78.6 | 82.3 | 83.5 | 80.2 | 66.4 | 91.8 | 3.9 | 10.1 |
| m · v1 rename s1 | 93.0 | 77.0 | 85.9 | 85.0 | 84.0 | 72.8 | 91.2 | 2.5 | 11.1 |
| m · v1 rename s2 | 92.7 | 76.1 | 81.8 | 84.6 | 84.9 | 68.1 | 92.2 | 3.0 | 10.8 |
| m · v1 s1 | 92.6 | 73.9 | 37.1 | 84.2 | 88.7 | 70.4 | 90.4 | 3.4 | 8.3 |
| m · v1 s2 | 92.3 | 74.2 | 44.8 | 85.7 | 84.9 | 58.8 | 92.1 | 3.7 | 8.7 |
| m · v1 s3 | 91.0 | 77.0 | 36.2 | 84.2 | 86.8 | 73.0 | 90.7 | 4.5 | 10.1 |
| soup · v1 · v1 s1 | 92.7 | 74.8 | 35.1 | 85.0 | 85.8 | 68.9 | 93.1 | 3.2 | 4.2 |
| soup · v1 · v1 s1 · v1 s2 | 93.1 | 74.8 | 37.7 | 85.3 | 86.8 | 64.8 | 93.1 | 2.7 | 5.4 |
| soup · v1 · v1 s1 · v1 s2 · v1 s3 | 92.5 | 75.2 | 37.3 | 84.6 | 87.7 | 66.7 | 91.7 | 3.2 | 6.1 |
| soup · v1 · v1 s1 · v1 s3 | 93.4 | 75.5 | 36.1 | 84.2 | 85.8 | 70.4 | 92.1 | 3.5 | 5.6 |
| soup · v1 · v1 s2 | 92.6 | 74.8 | 38.2 | 84.6 | 85.8 | 62.3 | 92.1 | 3.5 | 5.2 |
| soup · v1 · v1 s2 · v1 s3 | 92.7 | 74.8 | 38.1 | 84.2 | 86.8 | 64.9 | 92.2 | 3.2 | 5.3 |
| soup · v1 · v1 s3 | 92.2 | 76.4 | 35.2 | 83.5 | 85.8 | 70.1 | 92.1 | 3.5 | 4.9 |
| soup · v1 rename · v1 rename s1 | 93.8 | 78.6 | 85.5 | 83.9 | 82.1 | 68.7 | 92.5 | 2.6 | 9.3 |
| soup · v1 rename · v1 rename s2 | 93.3 | 77.3 | 83.6 | 83.9 | 84.0 | 68.2 | 93.6 | 3.1 | 9.9 |
| soup · v1 rename s1 · v1 rename s2 | 93.9 | 75.8 | 85.9 | 86.1 | 84.0 | 70.8 | 91.8 | 2.6 | 9.7 |
| soup · v1 s1 · v1 s2 | 93.0 | 75.2 | 40.2 | 85.7 | 86.8 | 63.8 | 93.0 | 3.1 | 7.6 |
| soup · v1 s1 · v1 s2 · v1 s3 | 92.6 | 75.5 | 38.3 | 84.6 | 88.7 | 67.1 | 91.7 | 3.1 | 7.7 |
| soup · v1 s1 · v1 s3 | 92.5 | 75.8 | 36.9 | 83.5 | 85.8 | 72.0 | 91.5 | 4.5 | 8.0 |
| soup · v1 s2 · v1 s3 | 91.9 | 74.2 | 40.1 | 83.5 | 86.8 | 65.1 | 91.1 | 3.5 | 8.9 |
| lerp · base · v1 · 0.1 | 79.1 | 62.1 | 30.3 | 82.8 | 70.8 | 78.8 | - | 31.2 | 21.3 |
| lerp · base · v1 · 0.2 | 82.1 | 64.6 | 30.9 | 83.9 | 76.4 | 77.4 | - | 29.4 | 18.1 |
| lerp · base · v1 · 0.3 | 85.1 | 68.3 | 32.0 | 83.9 | 78.3 | 77.2 | - | 26.6 | 14.4 |
| lerp · base · v1 · 0.5 | 89.6 | 71.7 | 32.8 | 83.9 | 84.9 | 75.3 | - | 16.8 | 7.3 |
| lerp · base · v1 · 0.7 | 91.9 | 71.7 | 33.4 | 83.2 | 84.9 | 71.4 | - | 6.3 | 4.4 |

| Model | Gen. | ML | Swap | HR | Inj. | Long | Order | ECE | Flip |
|---|---|---|---|---|---|---|---|---|---|
| lerp · base · v1 · 0.9 | 91.5 | 72.7 | 33.5 | 83.9 | 85.8 | 67.3 | - | 2.9 | 1.9 |
| lerp · base · v1 · 1.2 | 92.6 | 74.5 | 34.3 | 82.8 | 83.0 | 64.5 | - | 5.3 | 2.2 |
| lerp · base · v1 · 1.5 | 91.8 | 73.3 | 33.4 | 82.4 | 80.2 | 60.7 | - | 5.8 | 4.9 |
| lerp · base · v1 hr · 0.2 | 82.4 | 66.8 | 33.1 | 86.4 | 75.5 | 80.3 | - | 30.7 | 19.6 |
| lerp · base · v1 hr · 0.4 | 86.9 | 70.5 | 36.4 | 87.5 | 81.1 | 79.9 | - | 21.5 | 14.6 |
| lerp · base · v1 hr · 0.6 | 90.7 | 75.2 | 39.9 | 89.4 | 84.9 | 78.3 | - | 7.7 | 12.0 |
| lerp · base · v1 rename · 0.2 | 83.0 | 66.1 | 44.6 | 85.3 | 76.4 | 79.2 | - | 30.8 | 18.2 |
| lerp · base · v1 rename · 0.4 | 88.0 | 72.7 | 60.0 | 85.0 | 82.1 | 77.2 | - | 21.4 | 12.7 |
| lerp · base · v1 rename · 0.6 | 90.6 | 76.4 | 73.5 | 86.4 | 86.8 | 72.8 | - | 8.2 | 9.5 |
| lerp · base · v1 rename · 0.8 | 92.1 | 78.0 | 80.0 | 83.9 | 84.9 | 69.5 | - | 2.1 | 8.8 |
| lerp · base · v1 rename · 1.2 | 90.8 | 78.3 | 83.6 | 83.5 | 77.4 | 63.8 | - | 5.8 | 12.6 |
| lerp · v1 · a11 joint s0 · 0.5 | 93.5 | 77.3 | 69.4 | 91.9 | 92.5 | 89.0 | 91.9 | 4.1 | 5.7 |
| lerp · v1 · a11 joint s1 · 0.5 | 93.7 | 74.2 | 69.3 | 90.1 | 95.3 | 86.8 | 93.1 | 2.2 | 5.4 |
| lerp · v1 · v1 hr · 0.4 | 92.9 | 77.6 | 38.0 | 88.6 | 88.7 | 71.2 | - | 3.3 | 4.7 |
| lerp · v1 · v1 hr · 0.5 | 93.3 | 77.6 | 38.4 | 89.7 | 88.7 | 71.9 | - | 3.1 | 5.8 |
| lerp · v1 · v1 hr · 0.8 | 93.7 | 78.3 | 41.5 | 91.2 | 89.6 | 75.0 | - | 3.1 | 10.0 |
| lerp · v1 · v1 hr · 1.25 | 92.3 | 77.6 | 46.2 | 89.7 | 89.6 | 78.3 | - | 4.0 | 13.6 |
| lerp · v1 · v1 hr · 1.5 | 91.1 | 74.8 | 48.6 | 89.0 | 89.6 | 79.6 | - | 4.4 | 15.1 |
| lerp · v1 · v1 inject · 0.4 | 93.1 | 75.5 | 32.4 | 84.6 | 89.6 | 66.5 | - | 3.8 | 4.0 |
| lerp · v1 · v1 inject · 0.5 | 93.7 | 75.8 | 30.9 | 85.0 | 89.6 | 66.0 | - | 3.4 | 4.6 |
| lerp · v1 · v1 inject · 0.8 | 93.5 | 76.4 | 29.3 | 84.2 | 91.5 | 67.0 | - | 3.7 | 8.4 |
| lerp · v1 · v1 inject · 1.25 | 92.6 | 76.7 | 24.5 | 84.2 | 96.2 | 66.7 | - | 5.1 | 12.0 |
| lerp · v1 · v1 inject · 1.5 | 91.3 | 75.8 | 22.5 | 83.9 | 96.2 | 66.0 | - | 5.8 | 14.5 |
| lerp · v1 · v1 long · 0.4 | 92.6 | 77.0 | 34.8 | 83.9 | 87.7 | 83.2 | - | 3.5 | 4.3 |
| lerp · v1 · v1 long · 0.5 | 92.7 | 77.6 | 34.3 | 83.5 | 87.7 | 85.2 | - | 3.1 | 5.6 |
| lerp · v1 · v1 long · 0.8 | 93.0 | 77.6 | 35.4 | 83.9 | 87.7 | 90.7 | - | 3.4 | 9.5 |
| lerp · v1 · v1 long · 1.25 | 92.9 | 76.7 | 36.9 | 83.5 | 87.7 | 91.5 | - | 3.5 | 13.3 |
| lerp · v1 · v1 long · 1.5 | 91.7 | 74.8 | 38.6 | 83.2 | 87.7 | 92.6 | - | 4.7 | 15.1 |
| lerp · v1 · v1 noshuf · 0.4 | 93.0 | 75.2 | 40.4 | 83.5 | 85.8 | 66.5 | 93.0 | 3.7 | 4.0 |
| lerp · v1 · v1 noshuf · 0.5 | 92.7 | 75.5 | 41.9 | 83.9 | 85.8 | 66.0 | 92.7 | 3.5 | 4.9 |
| lerp · v1 · v1 noshuf · 0.8 | 92.3 | 75.5 | 46.4 | 83.2 | 84.9 | 65.1 | 90.3 | 3.8 | 7.2 |
| lerp · v1 · v1 rename · 0.2 | 92.9 | 75.2 | 50.7 | 84.2 | 87.7 | 68.7 | - | 3.3 | 2.5 |
| lerp · v1 · v1 rename · 0.4 | 93.0 | 75.5 | 64.7 | 85.3 | 86.8 | 69.7 | - | 3.2 | 3.9 |
| lerp · v1 · v1 rename · 0.5 | 93.5 | 77.3 | 70.9 | 83.9 | 86.8 | 70.1 | - | 3.6 | 5.2 |
| lerp · v1 · v1 rename · 0.6 | 93.1 | 78.6 | 75.8 | 84.6 | 86.8 | 69.8 | - | 3.7 | 6.2 |
| lerp · v1 · v1 rename · 0.8 | 93.3 | 79.8 | 80.2 | 83.9 | 84.0 | 68.2 | - | 3.8 | 8.1 |
| lerp · v1 · v1 rename · 1.25 | 91.0 | 77.0 | 82.4 | 85.0 | 77.4 | 63.5 | - | 5.6 | 12.6 |
| lerp · v1 · v1 rename · 1.5 | 88.6 | 75.5 | 82.0 | 83.9 | 73.6 | 59.1 | - | 6.9 | 15.5 |
| lerp · v1 · v1 s1 · 0.1 | 91.9 | 74.2 | 33.9 | 84.2 | 86.8 | 66.8 | 93.2 | 3.5 | 1.1 |
| lerp · v1 · v1 s1 · 0.2 | 92.3 | 73.6 | 34.3 | 84.6 | 87.7 | 67.3 | 93.8 | 4.1 | 2.5 |

| Model | Gen. | ML | Swap | HR | Inj. | Long | Order | ECE | Flip |
|---|---|---|---|---|---|---|---|---|---|
| lerp · v1 · v1 s1 · 0.3 | 92.9 | 74.5 | 35.4 | 85.0 | 86.8 | 68.1 | 93.4 | 3.6 | 3.6 |
| lerp · v1 · v1 s1 · 0.4 | 92.7 | 74.5 | 35.0 | 85.0 | 85.8 | 68.7 | 93.1 | 3.6 | 3.8 |
| lerp · v1 · v1 s1 · 0.5 | 92.7 | 74.8 | 35.1 | 85.0 | 85.8 | 68.9 | 93.1 | 3.2 | 4.2 |
| lerp · v1 · v1 s1 · 0.6 | 93.1 | 74.8 | 35.5 | 85.0 | 85.8 | 69.3 | 93.2 | 3.3 | 5.2 |
| lerp · v1 · v1 s1 · 0.7 | 93.0 | 74.5 | 36.2 | 84.6 | 85.8 | 69.8 | 92.6 | 3.2 | 5.7 |
| lerp · v1 · v1 s1 · 0.8 | 92.9 | 76.1 | 36.5 | 84.6 | 85.8 | 70.0 | 91.6 | 3.4 | 6.4 |
| lerp · v1 · v1 s1 · 0.9 | 92.6 | 75.2 | 37.1 | 85.0 | 86.8 | 70.3 | 91.7 | 3.2 | 6.9 |
| lerp · v1 rename · v1 rename s1 · 0.1 | 92.3 | 79.5 | 83.4 | 83.9 | 80.2 | 67.3 | 91.9 | 3.8 | 9.2 |
| lerp · v1 rename · v1 rename s1 · 0.2 | 93.0 | 79.8 | 84.6 | 83.2 | 83.0 | 67.6 | 92.6 | 3.6 | 9.2 |
| lerp · v1 rename · v1 rename s1 · 0.3 | 93.4 | 79.2 | 85.6 | 83.2 | 83.0 | 67.8 | 92.2 | 3.1 | 9.0 |
| lerp · v1 rename · v1 rename s1 · 0.4 | 93.9 | 78.9 | 85.4 | 83.5 | 82.1 | 68.6 | 92.6 | 3.5 | 9.0 |
| lerp · v1 rename · v1 rename s1 · 0.5 | 93.8 | 78.6 | 85.5 | 83.9 | 82.1 | 68.7 | 92.5 | 2.6 | 9.3 |
| lerp · v1 rename · v1 rename s1 · 0.6 | 94.1 | 77.6 | 86.0 | 84.6 | 82.1 | 70.0 | 92.5 | 2.5 | 10.1 |
| xfer · v1 s1 · v1 hr | 91.3 | 72.4 | 45.8 | 89.0 | 89.6 | 78.9 | - | 4.5 | 15.0 |
| xfer · v1 s1 · v1 inject | 92.5 | 73.9 | 27.9 | 82.8 | 94.3 | 69.2 | - | 5.3 | 13.8 |
| xfer · v1 s1 · v1 long | 91.5 | 70.8 | 41.5 | 82.4 | 89.6 | 92.6 | - | 4.9 | 14.6 |
| xfer · v1 s1 · v1 rename | 91.0 | 75.5 | 79.8 | 83.5 | 85.8 | 71.1 | - | 5.7 | 14.2 |
| xfer · v1 s2 · v1 hr | 90.7 | 72.0 | 49.2 | 89.4 | 88.7 | 68.7 | - | 5.8 | 14.9 |
| xfer · v1 s2 · v1 inject | 91.1 | 70.8 | 35.9 | 83.9 | 89.6 | 56.3 | - | 5.7 | 13.7 |
| xfer · v1 s2 · v1 long | 89.2 | 70.8 | 44.8 | 82.8 | 87.7 | 88.7 | - | 7.2 | 15.8 |
| xfer · v1 s2 · v1 rename | 88.4 | 74.5 | 80.5 | 86.1 | 81.1 | 55.8 | - | 6.4 | 15.5 |
| sum · hr · inject · 0.5 | 94.1 | 78.6 | 37.1 | 89.4 | 91.5 | 71.5 | - | 3.1 | 10.2 |
| sum · hr · inject · long · 0.333 | 94.3 | 78.3 | 37.4 | 88.3 | 90.6 | 83.6 | - | 2.5 | 9.8 |
| sum · hr · long · 0.5 | 93.8 | 78.3 | 40.9 | 89.0 | 88.7 | 89.3 | - | 3.0 | 10.8 |
| sum · inject · long · 0.5 | 94.1 | 77.3 | 33.1 | 83.5 | 92.5 | 85.5 | - | 3.3 | 8.5 |
| sum · rename · hr · 0.5 | 93.9 | 78.6 | 75.1 | 90.5 | 86.8 | 73.6 | - | 3.0 | 9.2 |
| sum · rename · hr · inject · 0.333 | 95.0 | 77.3 | 65.7 | 88.6 | 90.6 | 72.0 | - | 2.5 | 9.2 |
| sum · rename · hr · inject · long · 0.25 | 94.3 | 78.0 | 57.9 | 87.5 | 90.6 | 81.9 | - | 2.6 | 8.8 |
| sum · rename · hr · inject · long · 0.5 | 93.7 | 77.0 | 73.5 | 89.4 | 91.5 | 88.1 | - | 3.8 | 13.4 |
| sum · rename · hr · inject · long · 1.0 | 83.6 | 67.4 | 69.4 | 87.9 | 84.9 | 81.9 | - | 10.3 | 24.1 |
| sum · rename · hr · long · 0.333 | 94.2 | 78.3 | 64.2 | 89.0 | 88.7 | 85.4 | - | 2.8 | 8.8 |
| sum · rename · inject · 0.5 | 93.8 | 78.0 | 72.0 | 83.5 | 90.6 | 70.3 | - | 3.4 | 9.2 |
| sum · rename · inject · long · 0.333 | 93.8 | 79.2 | 62.3 | 82.8 | 90.6 | 82.7 | - | 3.1 | 8.4 |
| sum · rename · long · 0.5 | 93.0 | 78.9 | 71.8 | 82.8 | 85.8 | 86.3 | - | 3.8 | 9.8 |
| ties · baseroot · 0.2 · 1.0 | 93.8 | 78.9 | 67.1 | 88.6 | 89.6 | 84.1 | - | 4.3 | 10.9 |
| ties · v1root · 0.2 · 0.5 | 93.7 | 78.6 | 49.4 | 86.8 | 88.7 | 78.5 | - | 2.8 | 5.5 |
| ties · v1root · 0.2 · 1.0 | 94.1 | 77.3 | 64.7 | 88.3 | 91.5 | 85.5 | - | 3.2 | 11.0 |
| ties · v1root · 0.2 · 1.5 | 93.7 | 76.7 | 73.8 | 89.4 | 91.5 | 88.4 | - | 4.0 | 13.4 |
| ties · v1root · 0.5 · 1.0 | 94.1 | 76.1 | 68.2 | 88.6 | 91.5 | 85.5 | - | 3.4 | 11.5 |
| ties · v1root nosign · 1.0 · 1.0 | 94.5 | 77.3 | 63.0 | 87.5 | 91.5 | 84.1 | - | 2.7 | 10.4 |

| Model | Gen. | ML | Swap | HR | Inj. | Long | Order | ECE | Flip |
|---|---|---|---|---|---|---|---|---|---|
| ties · v1root trimonly · 0.2 · 1.0 | 93.1 | 79.2 | 46.6 | 86.4 | 88.7 | 76.6 | - | 3.0 | 5.2 |
| mask · hr · enc 0 13 | 93.4 | 78.6 | 36.1 | 90.8 | 88.7 | 72.5 | - | 2.7 | 10.8 |
| mask · hr · enc 14 27 | 93.3 | 74.2 | 44.1 | 85.3 | 84.9 | 71.4 | - | 4.3 | 2.7 |
| mask · hr · enc 21 27 | 92.5 | 73.9 | 32.9 | 84.2 | 86.8 | 66.5 | - | 3.9 | 0.5 |
| mask · hr · head | 92.6 | 73.6 | 32.8 | 84.2 | 86.8 | 66.2 | - | 3.8 | 0.1 |
| mask · inject · enc 0 13 | 93.7 | 77.0 | 27.3 | 84.6 | 90.6 | 67.0 | - | 2.8 | 9.6 |
| mask · inject · enc 14 27 | 92.2 | 74.5 | 32.4 | 83.5 | 88.7 | 66.8 | - | 4.0 | 2.1 |
| mask · inject · enc 21 27 | 92.6 | 74.2 | 32.6 | 84.2 | 86.8 | 66.4 | - | 4.4 | 0.6 |
| mask · inject · head | 92.6 | 73.6 | 32.6 | 84.2 | 86.8 | 66.0 | - | 4.8 | 0.0 |
| mask · long · enc 0 13 | 93.1 | 76.4 | 32.0 | 83.9 | 84.9 | 91.0 | - | 3.2 | 11.3 |
| mask · long · enc 14 27 | 92.6 | 74.2 | 35.8 | 83.9 | 88.7 | 71.2 | - | 2.8 | 2.6 |
| mask · long · enc 21 27 | 92.5 | 73.6 | 32.7 | 84.2 | 86.8 | 66.2 | - | 3.6 | 0.1 |
| mask · long · head | 92.6 | 73.6 | 33.0 | 84.2 | 86.8 | 66.2 | - | 5.4 | 0.1 |
| mask · rename · enc 0 13 | 92.6 | 78.0 | 71.2 | 84.6 | 84.9 | 69.7 | - | 4.6 | 8.8 |
| mask · rename · enc 14 27 | 93.1 | 74.8 | 68.5 | 82.8 | 87.7 | 69.7 | - | 4.3 | 2.9 |
| mask · rename · enc 21 27 | 92.6 | 73.9 | 35.3 | 83.9 | 86.8 | 67.1 | - | 4.0 | 0.4 |
| mask · rename · head | 92.6 | 73.6 | 32.8 | 84.2 | 86.8 | 66.2 | - | 2.5 | 0.0 |
| dare · 0.5 · 0.25 · 0 | 94.2 | 78.6 | 58.1 | 87.5 | 90.6 | 81.8 | - | 2.8 | 8.7 |
| dare · 0.5 · 0.25 · 1 | 94.3 | 77.6 | 57.9 | 87.9 | 90.6 | 81.9 | - | 2.6 | 9.1 |
| dare · 0.5 · 0.25 · 2 | 94.2 | 78.0 | 58.3 | 87.5 | 90.6 | 81.8 | - | 2.9 | 9.1 |
| dare · 0.5 · 1.0 · 0 | 83.3 | 67.1 | 69.8 | 86.8 | 85.8 | 82.1 | - | 10.2 | 24.5 |
| dare · 0.5 · 1.0 · 1 | 82.5 | 67.1 | 69.3 | 88.3 | 87.7 | 82.5 | - | 11.1 | 25.1 |
| dare · 0.5 · 1.0 · 2 | 83.0 | 67.7 | 69.4 | 87.2 | 84.9 | 81.9 | - | 10.3 | 24.6 |
| dare · 0.9 · 0.25 · 0 | 94.2 | 78.6 | 58.0 | 87.2 | 90.6 | 81.6 | - | 3.3 | 8.9 |
| dare · 0.9 · 0.25 · 1 | 93.9 | 78.9 | 57.7 | 87.9 | 90.6 | 81.9 | - | 3.2 | 8.9 |
| dare · 0.9 · 0.25 · 2 | 94.1 | 77.6 | 58.3 | 86.8 | 90.6 | 81.3 | - | 2.9 | 9.0 |
| dare · 0.9 · 1.0 · 0 | 83.3 | 66.1 | 69.4 | 87.9 | 82.1 | 81.3 | - | 9.5 | 25.0 |
| dare · 0.9 · 1.0 · 1 | 83.3 | 66.8 | 69.7 | 86.1 | 85.8 | 83.2 | - | 10.2 | 24.7 |
| dare · 0.9 · 1.0 · 2 | 82.8 | 66.1 | 69.1 | 87.5 | 84.9 | 81.0 | - | 9.9 | 25.3 |
| darenr · 0.9 · 1.0 · 0 | 92.7 | 75.8 | 43.0 | 85.3 | 88.7 | 73.1 | - | 3.3 | 3.7 |
| darenr · 0.9 · 1.0 · 1 | 92.9 | 75.8 | 43.2 | 85.3 | 88.7 | 73.4 | - | 3.3 | 3.6 |
| ens · v1 · v1 rename · v1 hr · v1 inject · v1 long | 93.9 | 81.1 | 41.2 | 86.1 | 89.6 | 72.8 | - | 1.5 | 6.2 |
| ens · v1 rename · v1 hr · v1 inject · v1 long | 94.6 | 80.1 | 44.8 | 85.0 | 90.6 | 74.2 | - | 1.5 | 8.6 |
| ensg · v1 · v1 rename · v1 hr · v1 inject · v1 long | 94.1 | 80.1 | 43.6 | 86.4 | 88.7 | 73.4 | - | 2.3 | 6.4 |
| ensg · v1 rename · v1 hr · v1 inject · v1 long | 94.9 | 80.1 | 46.4 | 86.8 | 90.6 | 74.7 | - | 2.2 | 8.7 |
| ensm · v1 · v1 rename · v1 hr · v1 inject · v1 long | 94.5 | 78.0 | 49.7 | 90.8 | 89.6 | 78.1 | - | 4.4 | 6.9 |
| ensm · v1 rename · v1 hr · v1 inject · v1 long | 95.0 | 78.6 | 50.5 | 90.8 | 89.6 | 78.1 | - | 4.0 | 8.7 |
| sumseed · v1 rename s1 · 0.25 | 94.3 | 78.9 | 58.6 | 87.9 | 90.6 | 83.0 | - | 2.9 | 8.9 |
| sumseed · v1 rename s1 · 1.0 | 82.5 | 66.5 | 71.8 | 86.8 | 89.6 | 86.2 | - | 9.3 | 24.8 |
| sumseed · v1 rename s2 · 0.25 | 93.8 | 77.0 | 58.0 | 87.5 | 88.7 | 81.8 | - | 3.0 | 8.6 |

| Model | Gen. | ML | Swap | HR | Inj. | Long | Order | ECE | Flip |
|---|---|---|---|---|---|---|---|---|---|
| sumseed · v1 rename s2 · 1.0 | 86.3 | 62.4 | 70.2 | 85.7 | 93.4 | 88.2 | - | 7.7 | 23.8 |

# Appendix B Weight geometry

**Table 22.** Distance between checkpoints: ||theta_a − theta_b|| (Frobenius norm of the difference, all parameters). The norm of the released weights is 1427.2 and the v1 fine-tune moved by 6.34.

| Model | v1 | s1 | s2 | s3 | ren | ren-s1 | hr | inj | long | nosh |
|---|---|---|---|---|---|---|---|---|---|---|
| v1 | - | 6.67 | 6.63 | 6.82 | 7.26 | 7.17 | 7.10 | 7.46 | 7.23 | 6.52 |
| s1 | 6.67 | - | 6.61 | 6.78 | 7.17 | 7.22 | 7.07 | 7.27 | 7.11 | 6.51 |
| s2 | 6.63 | 6.61 | - | 6.80 | 7.18 | 7.03 | 7.06 | 7.43 | 7.29 | 6.54 |
| s3 | 6.82 | 6.78 | 6.80 | - | 7.37 | 7.22 | 7.05 | 7.61 | 7.38 | 6.93 |
| ren | 7.26 | 7.17 | 7.18 | 7.37 | - | 7.41 | 7.55 | 7.67 | 7.50 | 7.20 |
| ren-s1 | 7.17 | 7.22 | 7.03 | 7.22 | 7.41 | - | 7.38 | 7.84 | 7.69 | 7.09 |
| hr | 7.10 | 7.07 | 7.06 | 7.05 | 7.55 | 7.38 | - | 7.82 | 7.47 | 7.11 |
| inj | 7.46 | 7.27 | 7.43 | 7.61 | 7.67 | 7.84 | 7.82 | - | 7.56 | 7.33 |
| long | 7.23 | 7.11 | 7.29 | 7.38 | 7.50 | 7.69 | 7.47 | 7.56 | - | 7.25 |
| nosh | 6.52 | 6.51 | 6.54 | 6.93 | 7.20 | 7.09 | 7.11 | 7.33 | 7.25 | - |

**Table 23.** Share of coordinates (%) that are bit-identical to the released weights after fine-tuning, in a few parts of the network (half-precision storage).

| Model | Embeddings | Head | Layer 0 | Layer 13 | Layer 27 |
|---|---|---|---|---|---|
| v1 | 81 | 2 | 6 | 7 | 7 |
| s1 | 81 | 2 | 6 | 7 | 7 |
| s2 | 81 | 2 | 6 | 7 | 7 |
| s3 | 81 | 2 | 6 | 7 | 7 |
| ren | 80 | 2 | 5 | 6 | 6 |
| ren-s1 | 80 | 2 | 5 | 6 | 7 |
| ren-s2 | 80 | 2 | 6 | 6 | 7 |
| hr | 80 | 2 | 6 | 7 | 7 |
| inj | 81 | 2 | 5 | 6 | 6 |
| long | 81 | 2 | 5 | 6 | 7 |
| nosh | 81 | 2 | 6 | 7 | 7 |
| nosh-s1 | 81 | 2 | 6 | 7 | 7 |
| nosh-s2 | 81 | 2 | 6 | 7 | 8 |